\pdfoutput=1
\documentclass[11pt]{article}

\usepackage{acl}

\usepackage{times}
\usepackage{latexsym}
\usepackage{multirow}
\usepackage{tabularx} 
\usepackage{ragged2e}
\usepackage{graphicx} 
\usepackage{array}
\usepackage{booktabs} 
\usepackage{geometry}      % Optional: For adjusting page margins
\usepackage{caption}       % Optional: For caption customization
\usepackage{subcaption}
\usepackage{amsmath}
\usepackage{comment}
\usepackage{calc} 
\usepackage[T1]{fontenc}
\usepackage[utf8]{inputenc}

\usepackage{microtype}

\usepackage{inconsolata}

\usepackage{graphicx}
\usepackage{arydshln}

\makeatletter
\def\adl@drawiv#1#2#3{%
        \hskip.5\tabcolsep
        \xleaders#3{#2.5\@tempdimb #1{1}#2.5\@tempdimb}%
                #2\z@ plus1fil minus1fil\relax
        \hskip.5\tabcolsep}
\newcommand{\cdashlinelr}[1]{%
  \noalign{\vskip\aboverulesep
           \global\let\@dashdrawstore\adl@draw
           \global\let\adl@draw\adl@drawiv}
  \cdashline{#1}
  \noalign{\global\let\adl@draw\@dashdrawstore
           \vskip\belowrulesep}}
\makeatother
\title{How Much Do LLMs Hallucinate across Languages? 
\\  On Realistic Multilingual Estimation of LLM Hallucination}
\author{%
  Saad Obaid ul Islam\textsuperscript{1} \quad
  Anne Lauscher\textsuperscript{2} \quad
  Goran Glavaš\textsuperscript{1} \\[1ex]
  \textsuperscript{1}WüNLP, CAIDAS, University of Würzburg \\ \texttt{\{saad.obaid-ul-islam,goran.glavas\}@uni-wuerzburg.de} \\
  \textsuperscript{2}Data Science Group, University of Hamburg \\ \texttt{anne.lauscher@uni-hamburg.de}
}

\newcommand{\rparagraph}[1]{\vspace{1.2mm}\noindent\textbf{#1}}

\newcommand{\sparagraph}[1]{\vspace{0.0mm}\noindent\textbf{#1}}

\begin{document}
\maketitle
\begin{abstract}
In the age of misinformation, hallucination---the tendency of Large Language Models (LLMs) to generate non-factual or unfaithful responses---represents the main risk for their global utility. Despite LLMs becoming increasingly multilingual, the vast majority of research on detecting and quantifying LLM hallucination are (a) English-centric and (b) focus on machine translation (MT) and summarization, tasks that are less common in realistic settings than open information seeking. 
%%%%%%%
In contrast, we aim to quantify the extent of LLM hallucination across languages in knowledge-intensive long-form question answering (LFQA). To this end, we train a multilingual hallucination detection model and conduct a large-scale study across 30 languages and 6 open-source LLM families. We start from an English hallucination detection dataset and rely on MT to translate-train a detection model. We also manually annotate gold data for five high-resource languages; we then demonstrate, for these languages, that the estimates of hallucination rates are similar between silver (LLM-generated) and gold test sets, validating the use of silver data for estimating hallucination rates for other languages.     
\textcolor{black}{For the final rates estimation, we build open-domain QA dataset for 30 languages with LLM-generated prompts and Wikipedia articles as references. Our analysis shows that LLMs, in absolute terms, hallucinate more tokens in high-resource languages due to longer responses, but that the actual hallucination rates (i.e., normalized for length) seems uncorrelated with the sizes of languages' digital footprints. We also find that smaller LLMs hallucinate more, and significantly, LLMs with broader language support display higher hallucination rates.} 
%estimated hallucination rates `in the wild' poorly correlate with models' language competency: while LLMs do generate longer   

%(1) LLMs generally hallucinate more in languages with higher digital representation and (2) larger models tend to hallucinate less than their smaller counterparts. 
%more in languages in which they are more fluent:  Our findings show that the ability of LLM to generate fluent text and hallucinations are highly correlated and on average, smaller models hallucinate less than larger models. 
\end{abstract}

\begin{figure*}[t]
    \centering
    \includegraphics[width=\textwidth]{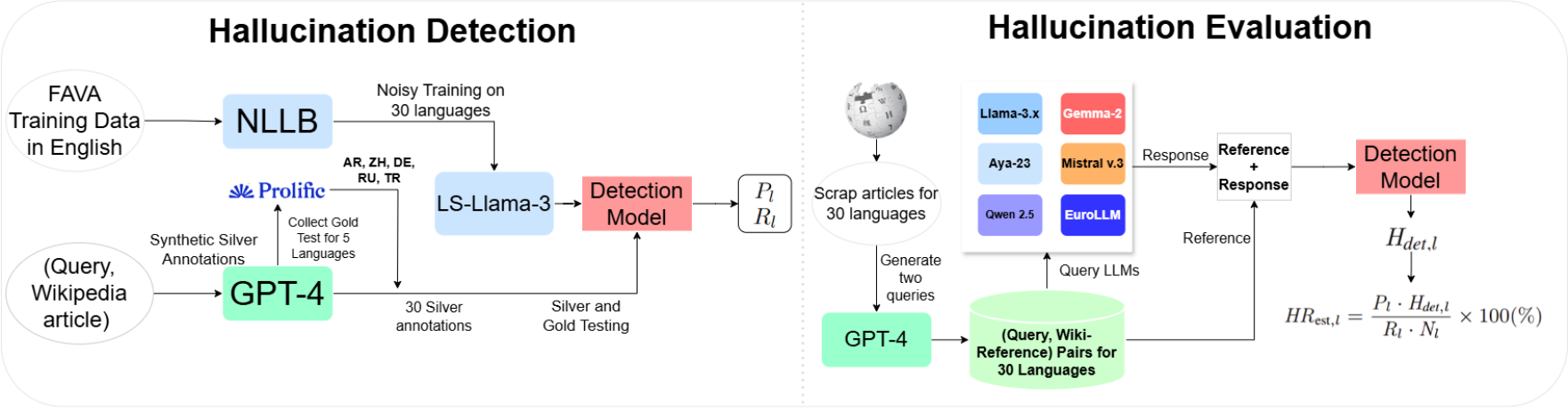}
    \caption{Illustration of our approach for estimating hallucination rates. \textbf{Hallucination Detection and Model Evaluation} (left side): (1) We automatically translate the English FAVA \cite{mishra2024finegrained} dataset to 30 languages and train our multilingual hallucination detection (HD) model on this (noisy) multilingual training data; (2) We synthesize a \textit{silver} multilingual hallucination evaluation dataset by prompting a state-of-the-art LLM (GPT-4) to introduce hallucinations in its answers to knowledge-seeking questions; for a subset of five high-resource languages, we additionally collect \textit{gold} (i.e., human) hallucination annotations; we dub this 30-language evaluation benchmark \textsc{mFAVA}. We use \textsc{mFAVA} to estimate HD model's per-language performances (precision and recall). \textbf{Hallucination Rate Estimation} (right side): (3) We estimate the hallucination rates for all 30 languages and six different LLM families from the number of detections of the HD model and its performance.             
    %, we first build large silver (through GPT-4) and gold test (through prolific) sets for 30 languages and train a detection model on noisy training data. For \textbf{Hallucination Evaluation} of LLMs, we scrap Wikipedia articles for 30 languages and create a synthetic (prompt,reference) dataset which we use to collect responses from 6 popular LLMs. Lastly, we detect hallucinations on the collected responses and adjust.
    }
    \label{fig:pipe-line}
\end{figure*}

\section{Introduction}\label{sec:intro}

Generalizing seamlessly to (seemingly) arbitrary language understanding, reasoning, and generation tasks, Large Language Models (LLMs) \citep{kojima2022large, dubey2024llama, aryabumi2024aya, yang2024qwen2} have arguably become the first ubiquitously adopted language technology, with application ranging from search engines \citep{xiong2024search}, interactive agents \citep{llminterativeagent} and knowledge retrieval \citep{yu2023generate} to various content generation tasks \citep{liu-etal-2022-multi}. Their utility, however, is hindered by their tendency to \textit{hallucinate} \citep{maynez-etal-2020-faithfulness,zhou2021detecting,ji2023survey, zhang2023siren}, that is, produce information that is either (i) inaccurate or factually incorrect with respect to the objective state of the world (e.g., in open-ended question answering) or (ii) unfaithful with respect to some reference (e.g., in summarization).

Consequently, a large body of work on tackling LLM hallucination has emerged, with efforts falling into the three main areas: (1) detection, i.e., identification of the hallucinated content; (2) evaluation, primarily focusing on measures for quantifying the extent and severity of hallucinations ; and (3) mitigation, focusing on mitigating hallucinative tendencies of LLMs \citep{ji2023survey}. While significant progress has been made in English \citep{maynez-etal-2020-faithfulness, liu-etal-2022-token, obaid-ul-islam-etal-2023-tackling, kasai2024realtime,  mishra2024finegrained}, hallucination evaluation efforts targeting other languages have been much sparser \cite{clark2023seahorse,guerreiro2023hallucinations,herrlein2024anhalten,shafayat2024multifact}.  
Moreover, these efforts have primarily targeted high-resource languages \citep{qiu-etal-2023-detecting, shafayat2024multifact} with benchmarks limited to reference-based tasks---text summarization \citep{clark2023seahorse, aharoni2022mface} and machine translation \citep{dale2023halomi, guerreiro2023hallucinations}. While highly relevant, these tasks are arguably less representative of LLM usage in realistic scenarios \cite{trippas2024users}, where knowledge-intensive long-form question answering (LFQA) is more prominent. 

In this work, we address the above gaps in multilingual hallucination detection and evaluation research with the ultimate goal of \textit{estimating the hallucination rates of LLMs across languages in a realistic scenario of open-domain QA}. Multilingual estimation of such hallucination rates is challenging due to the scarcity of multilingual hallucination benchmarks covering open-ended knowledge-seeking tasks that are representative of real-world LLM usage: unlike in reference-based generation tasks like summarization and machine translation, LLMs often generate long-form responses to open-ended questions, requiring more comprehensive evaluation approaches \citep{wei2024long}. Concretely, we present a large-scale study that estimates hallucination rates for 30 languages (both high(er)- and low(er)-resource languages). 
%%%
Our main contributions are as follows: 
\textcolor{black}{\textbf{(1)} We translate-train \cite{artetxe2023revisiting,ebing2024translate} a multilingual hallucination detection (HD)  model on 30 languages.
\textbf{(2)} We create \textsc{mFAVA} HD evaluation datasets with span-level human annotations (\textsc{mFAVA-gold}) for five high-resource languages, generate synthetic (\textsc{mFAVA-silver}) HD evaluation datasets for 25 additional languages, and validate the use of \textsc{mFAVA-silver} by showing the \textsc{mFAVA-silver} and \textsc{mFAVA-gold} estimates yield similar results; 
\textbf{(3)} We propose a protocol for estimating hallucination rates in open domain LFQA of LLMs and introduce an extensive synthetic dataset (51,133 prompts across 30 languages) for estimating LLM hallucination rates in highly multilingual settings;
\textbf{(4)} We offer a comprehensive hallucination rate analysis of six LLM families, validating previous findings that larger models tend to hallucinate less, and uncovering that broader LLM language coverage correlates with increased hallucination rates. 
This work is the first to estimate LLM hallucination rates for a wide range of languages using knowledge-intensive open-domain LFQA, reflecting real-world usage. Our comprehensive framework is illustrated in Figure~\ref{fig:pipe-line}\footnote{We release our datasets and work on: \url{https://github.com/WorldHellow/mHallucinations-LLM}}.}

%  multilingual setting for knowledge-intensive QA. An overview of our pipeline is provided in .

\section{Background and Related Work}
We provide a brief overview of the body of related work on (1) hallucination detection models and (2) benchmarks for evaluating LLM hallucination.

\rparagraph{Hallucination Detection.}
Coarsely, LLM hallucinations fall into two categories. \textit{Intrinsic} hallucination are content contradicts some reference information source. The reference may be explicitly given to the LLM as part of the task (e.g., the text to be summarized in summarization or source language text in machine translation) or it may implicit (e.g., general world knowledge in question answering). In contrast, \textit{extrinsic} hallucination refers to content that does not contradict the reference but is unnecessary or superfluous with respect to the task (e.g., additional facts in fact-based question answering) \citep{ji2023survey}. Recent work introduced finer-grained taxonomies for both categories. For example, \citet{mishra2024finegrained} distinguish between several types of intrinsic hallucinations (e.g., entity-based hallucinations or relation-based hallucinations). In a similar vein, extrinsic hallucinations are split into subtypes such as \textit{invented}, \textit{subjective}, and \textit{unverifiable} content. 

Unsurprisingly, most hallucination detection (and classifications) models are based on neural languages models. These are are either pre-trained encoder LM \citep{zhou2021detecting,liu-etal-2022-multi}, discriminatively fine-tuned to classify texts as containing hallucinations or not or LLMs prompted (zero-shot or with in-context examples) to detect hallucinations \citep{manakul-etal-2023-selfcheckgpt, yang2023new} or fine-tuned to generate hallucinated spans \citep{mishra2024finegrained}. 
In this work, we cast hallucination detection as a span-detection task, formulated discriminatively, with a classifier on top of an ``encoder-based'' LM. However, instead of resorting to small pretrained encoder LMs, we bidirectionally (i.e., discriminatively) fine-tune a larger generative LLM, following recent advances in converting decoder LMs into encoders \cite{li2023label, Dukic2024LookingRI, behnamghader2024llmvec, schmidt-etal-2024-self}.     

\rparagraph{Hallucination Benchmarks.}
Hallucination detection models as well as evaluation datasets have largely focused on English vary in the granularity from document-level \citep{yang2023new} of annotations/predictions, over passage- and sentence-level annotations \citep{zhou2021detecting, manakul-etal-2023-selfcheckgpt}, to fine-grained token- or span-level annotations \citep{liu-etal-2022-token, mishra2024finegrained}. Notable examples include SelfCheckGPT \citep{manakul-etal-2023-selfcheckgpt}, HaluEval \citep{li2023halueval}, and ScreenEval \citep{lattimer2023fast}, which measure hallucination detection rates in summarization and single-fact question answering. Multilingual benchmarks for evaluating hallucination detection models remain sparse and focus on reference-based tasks like machine translation \citep{dale2023halomi} and summarization \citep{qiu-etal-2023-detecting} which poorly represent the LLM usage in the wild.   

Faithfulness in reference-based tasks is complemented by truthfulness (i.e., factuality) in question answering. 
%Hallucination evaluation complements detection by measuring the faithfulness of generated responses against reference content. 
Most benchmarks, e.g., TruthFulQA \cite{lin2022truthfulqa}, RealtimeQA \cite{kasai2024realtime}, FreshQA \cite{vu2023freshllms}, and SimpleQA \cite{wei2024measuring} here are English-centric and cover only questions that require a simple single-factoid answer. 
%and are focus only on English, and test knowledge for a single factoid. 
LongFact \citep{wei2024long}, Factscore \citep{min2023factscore} and mFactScore \cite{kim-etal-2024-analysis} do test LLMs truthfulness in generating long and free-form answers. However, LongFact is an English-only benchmark, whereas Factscore and mFactscore, albeit multilingual, cover a very specific domain of biographic questions.

\begin{comment}
\section{Methodology}

To address the lack of concrete \textbf{hallucination detection} benchmarks for many languages, we extend the FAVABench pipeline \citep{mishra2024finegrained} to create multilingual silver and gold test datasets. This involves a two-step process: (1) generating passages from Wikipedia articles and (2) introducing errors to simulate six distinct hallucination types. For five diverse high-resource languages—Arabic (AR), Chinese (ZH), German (DE), Russian (RU), and Turkish (TR)—we collect gold annotations using the Prolific platform. To build our detection models, we adopt a translate-train approach \citep{hu2020xtreme, fang2021filter}, translating the FAVABench training set to 31 target languages. The resulting models are evaluated on both silver and gold test sets.
For \textbf{hallucination evaluation} we design a synthetic (prompt, wiki-reference) dataset to evaluate knowledge-intensive responses across 31 languages. Our experiments demonstrate that silver test datasets serve as a reliable proxy for gold datasets when computing hallucination rates in high-resource languages. Additionally, we observe that single-source and multi-source models achieve comparable performance in hallucination detection. For high-resource languages, training solely on the five gold languages is sufficient for detecting hallucinations in languages the model was not explicitly trained on, highlighting the cross-linguistic generalization capability. Lastly, we collect responses from 6 LLMs and detect hallucinations using our detection model.
\end{comment}

\begin{figure}[t]
    \centering
    \includegraphics[width=\columnwidth]{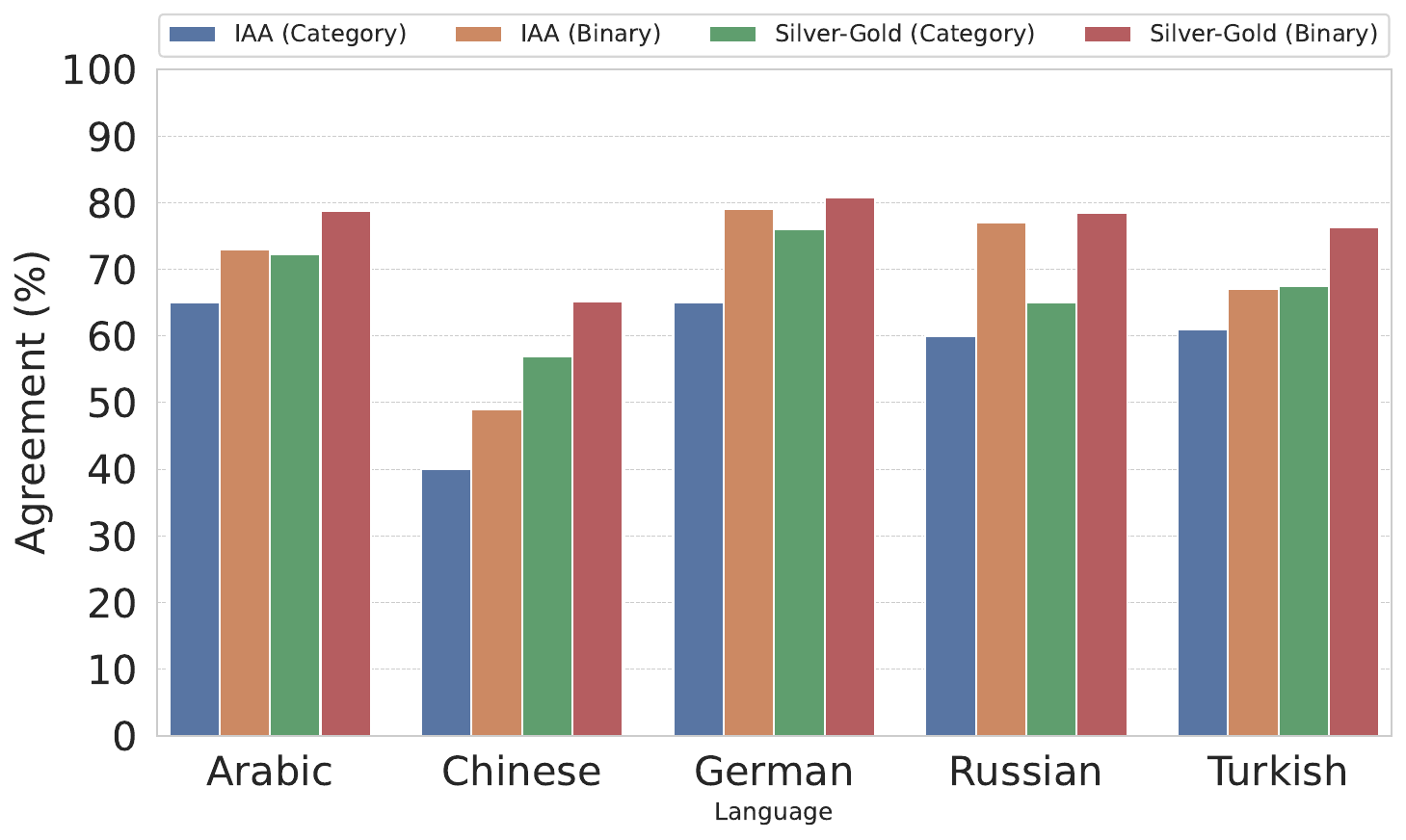}
    \caption{\textbf{1)} Inter-annotator agreement (IAA) for hallucination span detection (Binary; blue bars) and classification (Category; orange bars) for five high-resource languages; \textbf{2)} Hallucination span and class agreement between human labels and GPT-4 generated hallucinations (Silver-Gold; agreement on spans only: red bars; agreement on spans \textit{and} hallucination type: green bars).}
    \label{fig:agreement-metrics}
    \vspace{-0.5em}
\end{figure}

\section{Hallucination Detection}

We first describe how we obtained multilingual hallucination detection (HD) datasets (\S\ref{sec:mfava}) and then report on training and evaluation of a multilingual hallucination detection model (\S\ref{sec:model}). 

\subsection{\textsc{mFAVA} Benchmark}
\label{sec:mfava}

\rparagraph{HD Evaluation Datasets}. We start from the English FAVA \citep{mishra2024finegrained} dataset and its respective set of fine-grained hallucination types. FAVA's evaluation portions were created by (1) eliciting information-seeking prompts (i.e., questions) from various sources, (2) generating responses with three LLMs and (3) having human annotators label hallucinated span \footnote{See the original paper for more details and \S\ref{sec:appendix} for prompts for (2) and (3).}s. 
We follow a similar protocol to create evaluation datasets for 30 languages.\footnote{\S\ref{appendix:choice-languages} Figure \ref{tab:language_classification} lists the mFAVA languages.} We start from 300 information-seeking prompts, 150 from evaluation portion of FAVA and 150 from the Natural Questions dataset \citep{kwiatkowski-etal-2019-natural}. We then ask GPT-4 \cite{achiam2023gpt} to (1) first create answer passages in a target language and then to (2) explicitly introduce the hallucinations of the fine-grained FAVA types into the answer. We refer to these synthetically labeled hallucination evaluation datasets, comparable across the 30 target languages, as \textsc{mFAVA}-Silver. 

For five linguistically diverse high-resource languages---Arabic, Chinese, German, Russian, and Turkish---we also collect human hallucination annotations. To this end, we provide to the annotators the reference Wikipedia page, and the (hallucination-enriched) generation from \textsc{mFAVA}-Silver (of course, without the GPT-4's hallucination annotations). We source the annotations via Prolific, recruiting 5 annotators per language: all five annotators first annotated the same 50 instances, after which they were given non-overlapping sets of 50 more instances. We provide more details on the annotation process (and costs) in the \S\ref{appendix:annotation-process}. We measure the inter-annotator agreement (IAA) in terms of pairwise-averaged Cohen's kappa on token-level class decisions, both with (\textit{IAA Category}) and without (\textit{IAA Binary}) considering the fine-grained hallucination types. As shown in Figure \ref{fig:agreement-metrics}, we observe satisfactory to good IAA for all five languages. Regarding the 50 instances labeled by all annotators, we ultimately take the annotations of the annotator that has the highest IAA with hallucination annotatios of GPT-4 from \textsc{mFAVA}-Silver. We denote the final human-labeled evaluation datasets for the five high-resource languages with \textsc{mFAVA}-Gold. Figure \ref{fig:agreement-metrics} also shows the overall IAA between human annotations from \textsc{mFAVA}-Gold and GPT-4's synthetic annotations from \textsc{mFAVA}-Silver (\textit{Silver-Gold}): interestingly, we observe that human annotators on average agree more with GPT-4 than with one another. 

\begin{table}[t]
    \centering
    \resizebox{\columnwidth}{!}{%
    \renewcommand{\arraystretch}{1.1} % Slightly increase row height
    \small % Slightly increase font size
    \begin{tabular}{ccccccc}
        \toprule
        \textbf{Very Unlikely} & \textbf{Unlikely} & \textbf{Neutral} & \textbf{Likely} & \textbf{Very Likely} \\
        \midrule
         21.8\% & 24.7\% & 13.0\% & 25.3\% & 15.2\% \\
        \bottomrule
    \end{tabular}%
    }
    \caption{Annotator ratings for probability of augmented text fooling the reader for the 5 gold languages.}
    \label{tab:fool-rating}
    \vspace{-0.5em}
\end{table}

Because we synthesize the hallucinated content with GPT-4 (the annotators, of course, did not know that nor which part of the generation was meant to be a hallucination according to GPT-4), there is a risk that these hallucinations may not be realistic in the sense that they can fool a human reader. Because of this, we asked our annotators to additionally indicate (on a 5-degree Likert scale from ``very unlikely'' to ``very likely'') the likelihood of hallucination fooling a human reader for each span that they labeled. Table \ref{tab:fool-rating} reveals that more than half of the labeled hallucinations were judged as convincing (i.e., not unlikely to fool a human).  
The silver test set statistics for all 30 languages are shown in \S\ref{sec:appendix} Figure \ref{fig:silver-stacked-bar-chart}. Gold annotations statistics are shown in Table \ref{tab:gold-dataset-stats}.

\rparagraph{Training Dataset.} The FAVA training set, consisting of ca. 30K instances, is fully synthetically created in the same way as the test portion, just without the human annotation step. We automatically translate the training portion of the FAVA dataset using NLLB \citep{costa2022no} to our 30 target languages. After translation, we project the span-level annotations to token-level labels using the simple Inside-Out (I-O) scheme \citep{ramshaw-marcus-1995-text}\footnote{In preliminary experiments, we also tested the B-I-O scheme, but I-O led to better span detection performance.}. Like our evaluation benchmark \textsc{mFava}, we prepare training data for two tasks: (1) detecting hallucinated spans, regardless of hallucination type (\textit{Binary} task: tokens are classified as either part of a hallucinated span or not) and (2) detection and hallucination type classification (\textit{Category} task: 7-way classification, tokens classified into one of 6 FAVA hallucination types or as not part of a hallucinated span).

\begin{table}[t]
    \centering
    \resizebox{\columnwidth}{!}{%
    \renewcommand{\arraystretch}{1.1} % Slightly increase row height
    \small % Slightly increase font size
    \begin{tabular}{l|cccccc|c}
        \toprule
        \textbf & \textbf{ENT} & \textbf{REL} & \textbf{INV} & \textbf{CON} & \textbf{UNV} & \textbf{SUB} & \textbf{Total} \\
        \midrule
        RU & 184 & 65 & 188 & 287 & 211 & 153 & 1{,}088 \\
        AR & 144 & 10 & 171 & 123 & 150 & 69  & 667    \\
        ZH & 264 & 18 & 259 & 282 & 265 & 139 & 1{,}227 \\
        DE & 546 & 25 & 311 & 324 & 333 & 238 & 1{,}777 \\
        TR & 149 & 27 & 288 & 244 & 161 & 149 & 1{,}018 \\
        \midrule
        \textbf{Total} & 1{,}287 & 145 & 1{,}217 & 1{,}260 & 1{,}120 & 748 & 5{,}777 \\
        \bottomrule
    \end{tabular}%
    }
    \caption{Hallucinated span counts in the gold dataset across languages. ENT (Entity), REL (Relation), INV (Invented), CON (Contradictory), UNV (Unverifiable), SUB (Subjective).}
    \label{tab:gold-dataset-stats}
    \vspace{-0.5em}
\end{table}

    \begin{table*}[t]
    \centering
    \setlength{\tabcolsep}{6pt}
    {\small
    \begin{tabular}{lllcccccccccc}
    \toprule
    & & &  \multicolumn{2}{c}{\textbf{German}} & \multicolumn{2}{c}{\textbf{Chinese}} & \multicolumn{2}{c}{\textbf{Arabic}} & \multicolumn{2}{c}{\textbf{Russian}} & \multicolumn{2}{c}{\textbf{Turkish}} \\
    \cmidrule(lr){4-5} \cmidrule(lr){6-7} \cmidrule(lr){8-9} \cmidrule(lr){10-11} \cmidrule(lr){12-13}
                        \textbf{Task} & \textbf{Model} & \textbf{Context} & \textbf{Silver} & \textbf{Gold} & \textbf{Silver} & \textbf{Gold} & \textbf{Silver} & \textbf{Gold} & \textbf{Silver} & \textbf{Gold} & \textbf{Silver} & \textbf{Gold} \\ 
    \midrule

                  & \textsc{Mono} & \textit{Bidirect} & 78.0 & 58.0 & 62.4 & 55.1 & 75.3 & 54.4 & 78.9 & 60.7 & 78.5 & 66.7 \\ 
    Binary                    & \textsc{Multi} & \textit{Bidirect} & \textbf{89.5*} & \textbf{65.0*} & 69.7 & 58.7 & \textbf{82.5*} & \textbf{61.6} & \textbf{89.1*} & \textbf{65.5*} & \textbf{86.4*} & \textbf{72.5*} \\ \cdashlinelr{2-13}
                        & \textsc{Multi} & \textit{Causal}        & 81.8 & 59.6 & \textbf{76.3*} & \textbf{62.2*} & 75.3 & 60.0 & 75.8 & 55.6 & 75.7 & 67.3 \\
\midrule
                         
                & \textsc{Mono} & \textit{Bidirect}  & 53.4 & 38.3 & 35.2 & 22.6 & 14.6 & 7.3  & 63.3 & 36.2 & 49.1 & 30.3 \\ 
    Category                    & \textsc{Multi} & \textit{Bidirect}   & \textbf{73.2*} & \textbf{45.0} & 46.5 & 30.1 & \textbf{66.1*} & \textbf{37.2*} & \textbf{72.3*} & \textbf{41.5*} & \textbf{72.9*} & \textbf{51.8*} \\ \cdashlinelr{2-13} 
                        & \textsc{Multi} & \textit{Causal} & 68.7 & 43.4 & \textbf{56.5*} & \textbf{34.1*} & 51.8 & 29.4 & 62.6 & 37.9 & 58.6 & 42.4 \\
    \bottomrule
    \end{tabular}}
    \caption{Token-level F1 performance of multilingual (\textsc{Multi}) and monolingual (\textsc{Mono}) hallucination detection models for five high-resource languages with both \textit{Silver} and \textit{Gold} evaluation data in \textsc{mFAVA}. Performance reported for hallucination detection alone (\textit{Binary}) and hallucination detection and type classification (\textit{Category}). Models fine-tuned without (\textit{Bidirect}) or with (\textit{Causal}) future token masking. \textbf{Bold}: best result in each column, \textbf{Asterisk}: significantly higher ($p<0.05$) score.}
    \label{tab:detection-training-results}
    \vspace{-0.5em}
    \end{table*}

\subsection{Multilingual Hallucination Detection}
\label{sec:model}

\rparagraph{Models.} Using the translations of ca. 30K FAVA training instances in our 30 target languages, we train the following models: (1) \textsc{Mono} denotes monolingual models trained on data of one language (and evaluated for the same language on the respective \textsc{mFAVA} portion), i.e., we train 30 \textsc{Mono} models, one for each of our target languages; (2) \textsc{Multi} refers to a single multilingual model trained on concatenated training data of all 30 languages. We train the Mono models and the Multi model for both tasks, \textit{Binary} and \textit{Category}. We follow the recent body of work that successfully converts generative decoder LLMs into encoders for discriminative tasks \cite{li2023label, Dukic2024LookingRI, behnamghader2024llmvec, schmidt-etal-2024-self} and fine-tune Llama-3-8B-base \cite{dubey2024llama} by removing future-token masking, i.e., allowing for bidirectional contextualization (\textit{Bidirect}). For comparison, for the \textit{Multi} model, we also fine-tune the decoder as-is, using the default causal token masking (i.e., unidirectional contextualization; \textit{Causal}). 

\rparagraph{Training.} 
In all cases, we freeze the original model parameters and train QLora adapters \citep{dettmers2024qlora}, with three runs (random seeds) for each experiment, reporting mean performance. The input to the models is the reference Wikipedia article, prepended to the LLM-generated answer, with the cross-entropy loss computed exclusively over the tokens of the LLM-generated answer. We provide further training details in \S\ref{appendix:training-details}.        

\rparagraph{Results.}
Table \ref{tab:detection-training-results} summarizes the hallucination detection performance for five high-resource languages for which we have both LLM-synthesized Silver data and human-annotated Gold portions in our \textsc{mFAVA} benchmark. 
We first observe that, expectedly, just detecting hallucinated spans (\textit{Binary} task) is much easier than additionally correctly recognizing the type of hallucination (\textit{Category} task). 
%expectedly, much easier Straightaway, we can observe that binary labels consistently yield better results across all models and languages. 
Although category labels offer finer-grained insight into the nature of LLM hallucination, we deem the models' performance on fine-grained hallucination type classification---especially on Gold, human-labeled portions of \textsc{mFAVA}---insufficient for reliably estimating type-specific hallucination rates (see \S\ref{sec:final-estimates}). These results are in line with IAA from Figure \ref{fig:agreement-metrics}, with consistently larger IAA for hallucination detection (Binary) then for type classification (Category). This renders fine-grained hallucination type classification difficult for both humans and models and warrants a broader research effort on hallucination type taxonomies as well as better hallucination type detection models. We leave this for future work. 

Models' performance on the detection-only (Binary) tasks is much better across the board, but the results are much better on the Silver portions (hallucinations generated by GPT-4) of \textsc{mFAVA} than on the Gold (human-labeled hallucination spans). This is expected, because the hallucinated spans in our training data have also been generated by GPT-4---this means that the human-annotated Gold \textit{mFAVA} portions introduce much more of a distribution shift w.r.t. training data than the corresponding Silver portions. At this point it is important to (re-)emphasize that we are not really interested in the absolute performance of the detection models, but rather using these detection performance estimates to produce reliable hallucination rate estimates for LLMs in realistic setting (\S\ref{sec:final-estimates}). 

We next observe that the 30-language multilingual model (\textsc{Multi}) is consistently better than language-specific monolingual models (\textsc{Mono}), with gaps being particularly wide in the Category task (e.g., +30\,F1 points for Arabic on the Gold \textsc{mFAVA} portion). Albeit smaller, the differences are also substantial in the Binary hallucination detection (e.g., +7\,F1 points for Arabic and German, on respective Gold \textsc{mFAVA} portions). Finally, bidirectional contextualization in fine-tuning (\textit{Bidirect}) seems to be generally more effective than fine-tuning with future-token masking (\textit{Causal}), with Chinese performance as the only exception. This is in line with findings from other token-classification tasks \cite{li2023label,Dukic2024LookingRI}.

\begin{figure*}[t]
    \centering
    \includegraphics[width=\textwidth]{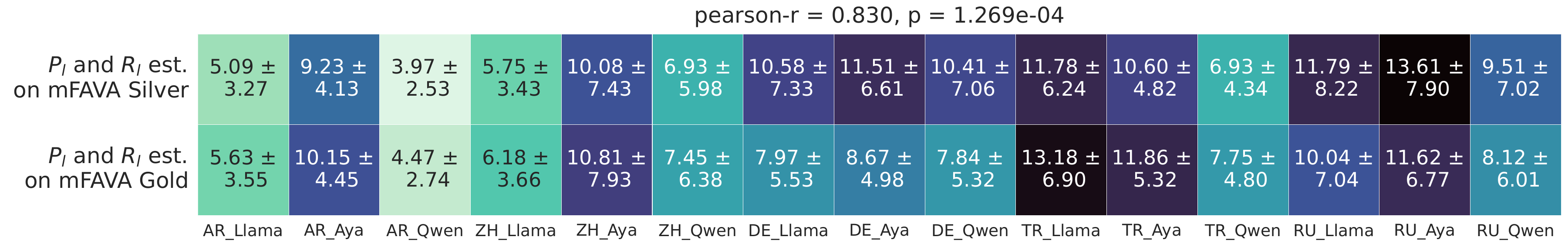}
    \caption{Comparison of hallucination rate estimates $\mathit{HR}_{\text{est}, l}$ (mean $\pm$ std over five LLM runs) for Arabic (AR), Chinese (ZH), German (DE), Russian (RU), and Turkish (TR) for 3 LLMs based on the estimates of $\mathit{P}_l$ and $\mathit{R}_l$ of the \textsc{Multi} (\textit{Bidirect}) model on (1) \textsc{mFAVA}-Silver (top row) and (2) \textsc{mFAVA}-Gold (bottom row). The two sets of estimates are highly correlated $(r=0.83, p=1.26e-04)$.}
    \label{fig:gold-heatmap}
\end{figure*}

\begin{figure*}[t]
    \centering
    \includegraphics[width=\textwidth]{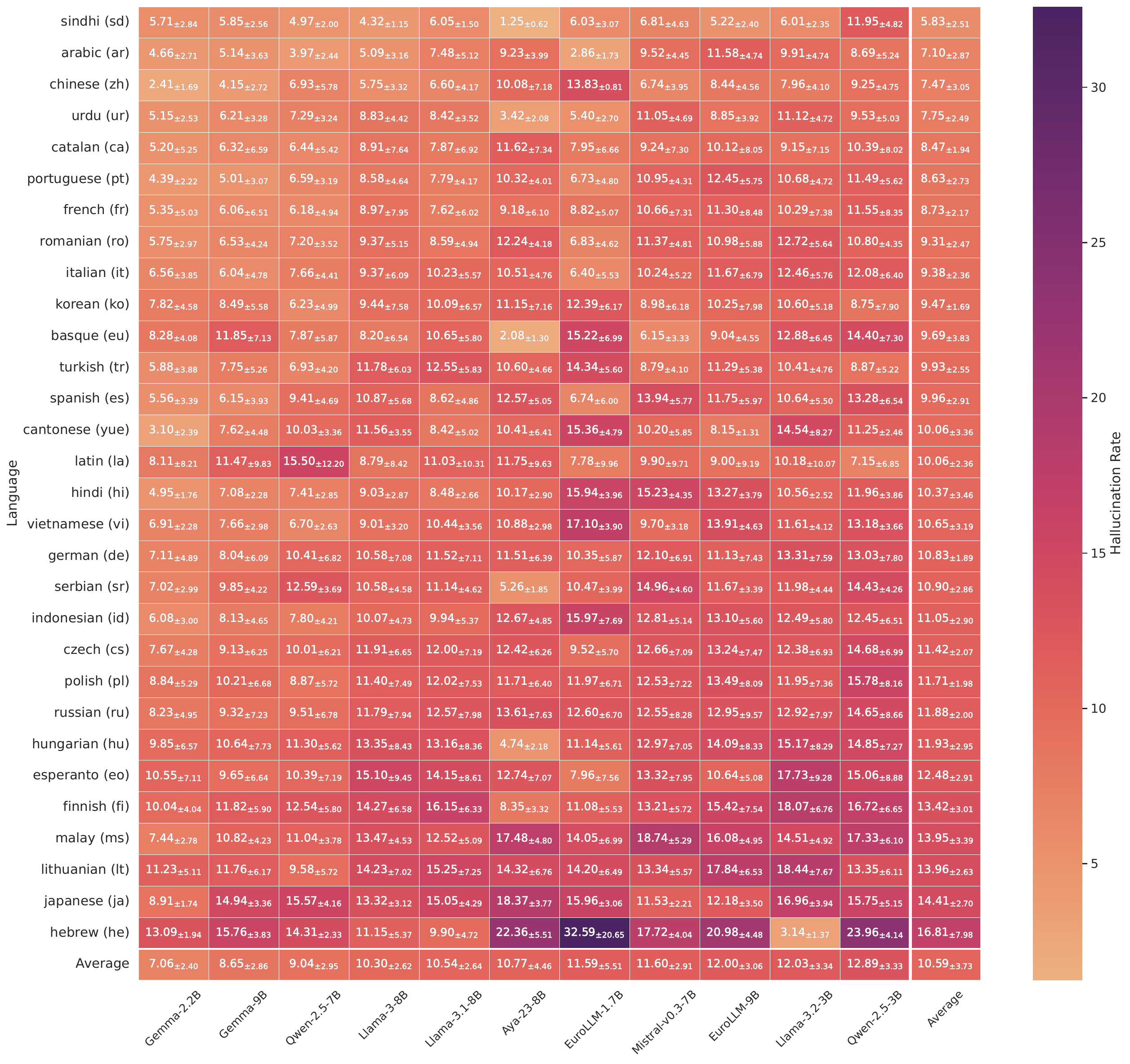}
    \caption{Mean estimates of in-the-wild hallucination rates ($\pm$ std) for 30 languages and 11 LLMs. Each mean score is an average of 15 $\mathit{HR}_{\text{est}, l}$ estimates, (3 different HD model instances applied to 5 different LLM responses). Average rates increase from top to bottom (over languages) and from left to right (over LLMs).}
    \label{fig:hallucination_rates_all}
    \vspace{-0.5em}
\end{figure*}

\section{Estimating Hallucination in Realistic Setting}
\label{sec:estimation_methodology}

We next propose a protocol for estimating hallucination rates of LLMs (for a wide range of languages), based (1) on the number of hallucinated tokens detected by a hallucination detection (HD) model in the wild and (2) estimates of HD model's performance (precision and recall).  

\subsection{From Model Performance to Hallucination Rates Estimates}
%%% GG: this was completely wrong, we don't apply this formula to balance out P & R
%For estimating hallucination rates, it is desirable for the detection model to have high and balanced precision and recall to ensure hallucinations are correctly identified without false positives. However, our detection system exhibits imbalanced precision and recall \footnote{See Appendix Table~\ref{tab:hallucination-metrics-extended}}. To ensure the reliability of hallucination rate estimates, we adjust for the detection system's precision and recall using the following formula 

\sparagraph{Estimating Hallucination Rates.} Let $\mathit{P}_l$ and $\mathit{R}_l$ be the estimates of token-level precision and recall of a HD model for some language $l$ and let $\mathit{H}_{\textit{det}, l}$ be the number of hallucination tokens that the HD model detected (i.e., predicted) on some corpus $C_l$ of LLM generations in language $l$, which serves as an approximation of the LLM outputs in the wild. We then posit that the estimate of the true hallucination rate of the LLM in the wild for language $l$, $\mathit{HR}_{\text{est}, l}$, is given as follows: 

\begin{small}
\label{equation:adjusted-formula}
\begin{equation}\label{eq:HR_adjusted}
\mathit{HR}_{\text{est}, l} = \frac{\mathit{P}_l \cdot \mathit{H}_{\textit{det}, l}}{\mathit{R}_l \cdot \mathit{N}_l} \times 100 (\%)
\end{equation}    
\end{small}

% See Appendix \ref{appendix:derivation} for derivation
% obviously, this holds if the in the wild corpus 

\noindent where $\mathit{N}_l$ is the total number of tokens in $C_l$, i.e., the total number of tokens generated by the LLM across answers to all user prompts. Intuitively, multiplying the number of model's detections $\mathit{H}_{\textit{det}, l}$ with its estimated precision $\mathit{P}_l$ discounts $\mathit{H}_{\textit{det}, l}$ by the number of tokens falsely detected as hallucinated by the model---while we do not know exactly which token predictions are false positives, the expected rate of false positives is, by definition, exactly captured by $\mathit{P}_l$. Analogously, dividing $\mathit{H}_{\textit{det}, l}$ with $\mathit{R}_l$ accounts for the tokens that are hallucinated, but will (falsely) not be detected by the model---and $\mathit{R}_l$ is exactly the estimate of the rate of such false negatives. We divide the estimate of the absolute number of truly hallucinated tokens (i.e., $\mathit{P}_l \cdot \mathit{H}_{\textit{det}, l} / \mathit{R}_l$) with $\mathit{N}_l$, making $\mathit{HR}_{\text{est}, l}$ a relative measure, that is, a rate (i.e., proportion) of all generated tokens that are hallucinated (multiplied by 100 and expressed as \%). 
%Finally, we multiply the rate by 100 to report it as percentage.% 
We provide a more detailed explanation/justification of Eq.\,\eqref{eq:HR_adjusted} in \S\ref{appendix:derivation}.   

%where for a language $l$ $\mathit{P}_l$ and $\mathit{R}_l$ are precision and recall of the model. $\mathit{H}_{\text{detected}, l}$ are the hallucinated tokens detected by the model, $\mathit{N}_l$ are the total generated tokens and $\mathit{HR}_{\text{adjusted}, l}$ is the adjusted rate. 

\rparagraph{Estimation Dataset.} We next create corpora $C_l$ (one corpus for each of our 30 target languages) of free-text LLM answers to knowledge-intensive queries, as approximations of the LLM usage in the real world. 
%There are currently no multilingual evaluation dataset designed to test LLMs on generating knowledge-intensive responses. Existing works, such as Factscore and mFactScore \citep{min2023factscore, kim-etal-2024-analysis} and LongFact \citep{wei2024long}, are limited in scope. mFactScore focuses on Wikipedia biography knowledge in only three languages, while Factscore and LongFact is centered on English. 
%%%
We start by randomly selecting articles from the language-specific Wikipedia, to serve as ground truth reference text. To ensure quality of reference text, we choose only from Wikipedia articles that are at least 2,000 characters long and have the collaborative Wikipedia depth \citep{alshahrani-etal-2023-depth} of at least 5.\footnote{The depth indicates the number of collaborative edits and correlates with the quality/factuality of the content.} 
%To address this, we constructed a synthetic prompt dataset using Wikipedia as the ground truth. We scraped Wikipedia articles for each language, ensuring each article had at least 2,000 characters and a Wikipedia depth \(\geq 5\) \footnote{Wikipedia depth is rough indicator of Wikipedia's collaborative quality. Higher depth is better}. 
We then prompt GPT-4 to generate two knowledge-intensive queries for each selected article, ensuring that the information required to answer to the query is fully contained in the article text (see Table \ref{tab:prompt} in the \S\ref{appendix:evaluation-dataset} for the exact prompt). As a sanity check, we manually checked for 50 synthesized queries and five languages from Table \ref{tab:detection-training-results}---by translating the query and reference article to English---whether the answers to queries are indeed contained in the article, establishing that this is indeed so in 98\% of cases.
%%%
Our final dataset for multilingual hallucination rate estimation consists of 25,685 Wikipedia articles (spanning over 15,940 unique Wikipedia categories) and 51,133 queries. Table \ref{tab:hallucination-eval-dataset} in \S\ref{sec:appendix} provides per-language statistics. We provide details on constructing the datasets in \S\ref{appendix:evaluation-dataset}.  

Finally, we collected responses to all queries from a total of 11 instruction-tuned open-source LLMs from 6 families (ranging in parameter count from ~2 to 9 billion): Llama-3.x \citep{dubey2024llama}, Aya-23 \citep{aryabumi2024aya}, Euro-LLM \citep{martins2024eurollm}, Gemma-2 \citep{gemma_2024} Qwen-2.5 \citep{yang2024qwen2}, and Mistral v3 \citep{jiang2023mistral}. We divided the queries into five subsets: for each subset the LLMs generated responses with a different random seed (see Table \ref{tab:generation-config} for details on the generation configurations).

\rparagraph{Estimates from \textsc{mFAVA}-Silver Performance.} 
\label{subsec:silver-vs-gold} 
On the one hand, creating Gold datasets for hallucination detection evaluation is prohibitively expensive (see \S\ref{appendix:annotation-process})---this is why we obtained such annotations for only five of 30 \textsc{mFAVA} languages. On the other hand, the estimates of HD model's performance are much higher on \textsc{mFAVA}-Silver (see Table \ref{tab:detection-training-results}), with GPT-4-labeled hallucinations: this, at first glance, questions the validity of estimating realistic hallucination rates based on $\mathit{P}_l$ and $\mathit{R}_l$ estimated on Silver data, for the 25 languages for which we do not have \textsc{mFAVA}-Gold portions. 
Recall, however, that we do not care about HD model's absolute $\mathit{P}_l$ and $\mathit{R}_l$, but whether the $\mathit{P}_l$ and $\mathit{R}_l$ estimates can produce reliable hallucination rate estimates $\mathit{HR}_{\text{est}, l}$. Looking at Eq.\,\ref{eq:HR_adjusted}, $\mathit{HR}_{\text{est}, l}$ depends on the ratio $\mathit{P}_l/\mathit{R}_l$ and not absolute  values of $\mathit{P}_l$ and $\mathit{R}_l$. 
%%%
We thus next test, for the five languages with both Silver and Gold portions in \textsc{mFAVA}, whether the $\mathit{HR}_{\text{est}, l}$ estimates based on the Silver $\mathit{P}_l$ and $\mathit{R}_l$ (roughly) match those based on Gold $\mathit{P}_l$ and $\mathit{R}_l$. Figure \ref{fig:gold-heatmap} shows $\mathit{HR}_{\text{est}, l}$ estimates, computed from the performance of our \textsc{Multi} (\textit{Bidirect}) model on Silver and Gold portions, respectively, and number of its hallucination detections $\mathit{H}_{\textit{det}, l}$  on outputs of three LLMs: Llama-3-8B, Qwen-2.5-7B, and Aya-8B. We observe very strong Pearson correlation $(r=0.83, p=1.26e-04)$ between the Gold-based and Silver-based $\mathit{HR}_{\text{est}, l}$ estimates, which, we argue, justifies the usage of Silver \textsc{mFAVA} datasets for estimating $\mathit{HR}_{\text{est}, l}$ for the 25 languages without the Gold \textsc{mFAVA} portions.

\begin{figure*}[t!]
    \centering
    % First subfigure
    \begin{subfigure}[t]{0.32\textwidth}
        \centering
        \includegraphics[width=\linewidth]{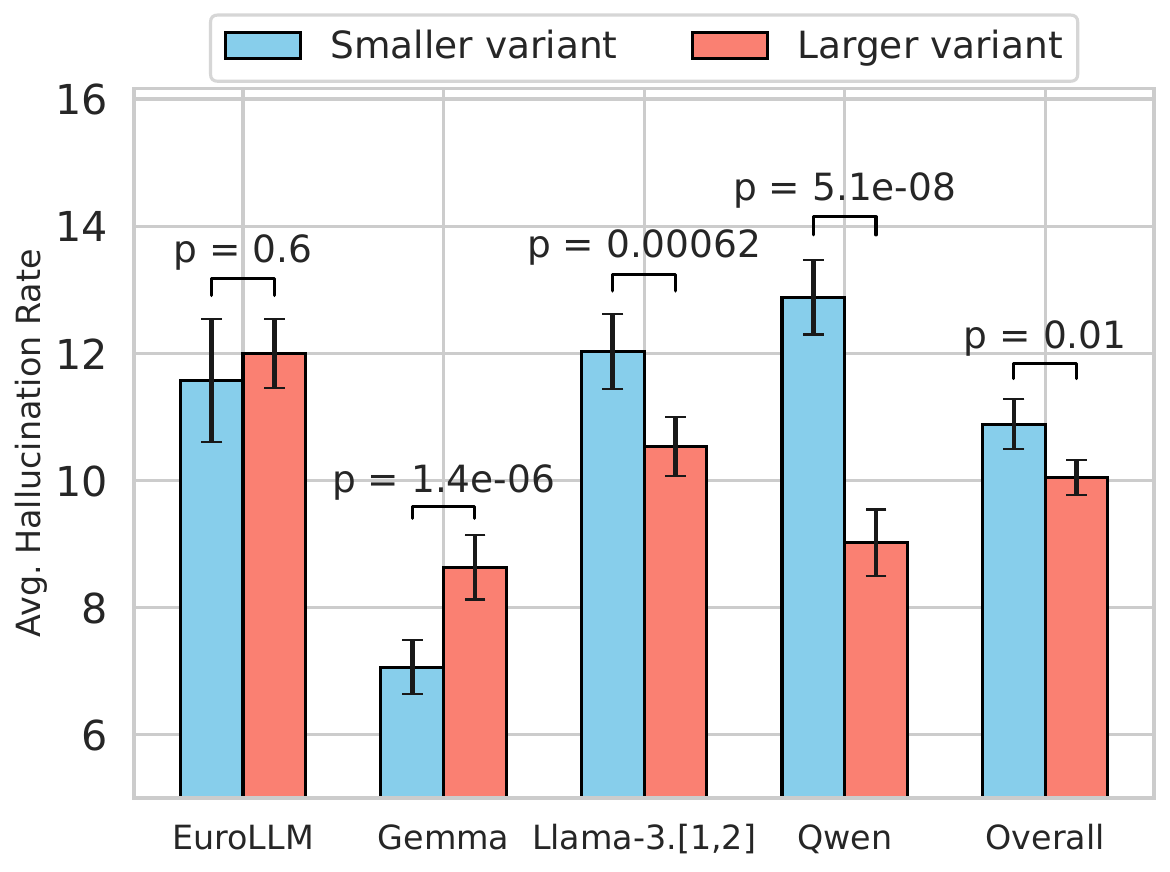}
        \caption{}
        \label{fig:paired-bar-graph}
    \end{subfigure}
    \hfill
    % Second subfigure
    \begin{subfigure}[t]{0.32\textwidth}
        \centering
        \includegraphics[width=\linewidth]{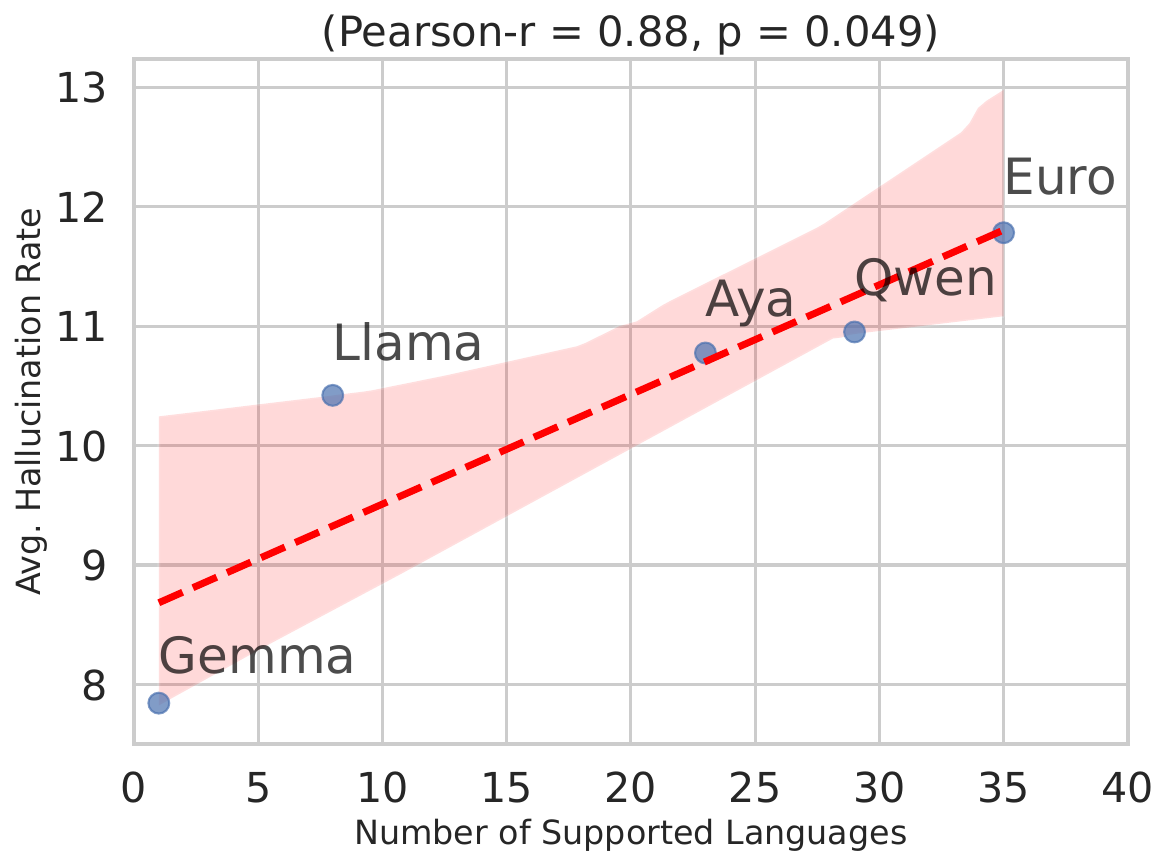}
        \caption{}
        \label{fig:rate_vs_supported_language}
    \end{subfigure}
    \hfill
    % Third subfigure
    \begin{subfigure}[t]{0.32\textwidth}
        \centering
        \includegraphics[width=\linewidth]{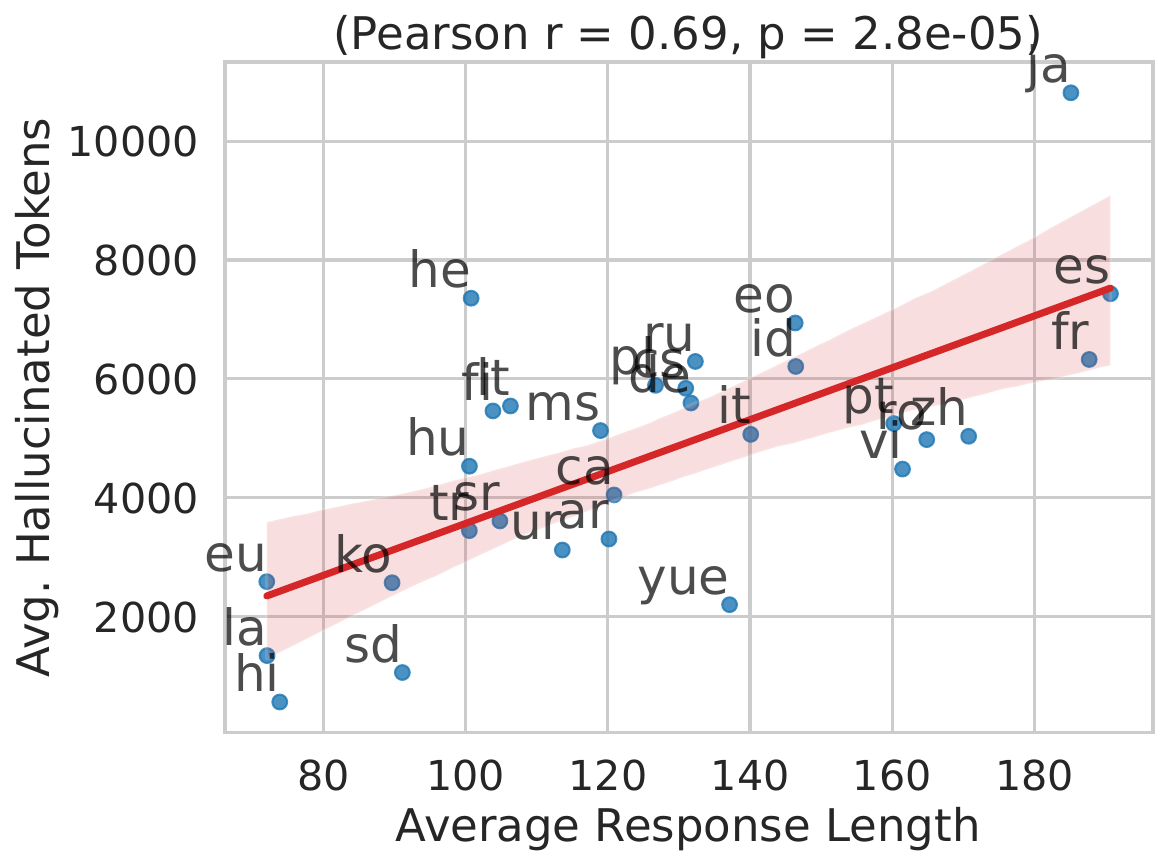}
        \caption{}
        \label{fig:overall}
    \end{subfigure}
    \label{fig:analysis}
    \caption{\ref{fig:paired-bar-graph} Larger models hallucinate significantly less than smaller ones. Bars are labeled with $p$-values from $t$-test. \ref{fig:rate_vs_supported_language} Correlation between hallucination rates (averaged over all 30 languages) and the officially declared number of supported languages. \ref{fig:overall} On average, as response length increases, so do the absolute hallucinations $\mathit{H}_{\text{detected}, l}$.}
\end{figure*}

\subsection{Final Estimates} \label{sec:final-estimates}

%Having shown that silver $\mathit{PR_l}$ and $\mathit{R_l}$ are a solid proxy for gold (cf. \S\ref{subsec:silver-vs-gold}), 
Figure \ref{fig:hallucination_rates_all} shows our in-the-wild hallucination rate estimates $\mathit{HR}_{\text{est}, l}$ for all 30 \textsc{mFAVA} languages and 11 LLMs. The average rate across all languages varies between 7\% and 12\%, with both Gemma models offering the lowest rates. Smaller Qwen-2.5 (3B) model hallucinates the most and, interestingly, significantly more than its larger counterpart (9B). 
%On average, Gemma-2B hallucinates the least (maximum rate $13.03\%$ for Hebrew) while Qwen-2.5-3B hallucinates the most (maximum rate $23.96\%$ for Hebrew). 

\rparagraph{More Parameters, Less Hallucination?} %For four LLMs families---Gemma, Qwen, EuroLLM, and Llama---we benchmarked models of different sizes. We next test whether smaller variants hallucinate significantly more or less than larger counterparts. 
%the differences in  We next the significance of difference in hallucination rates for different models sizes  Overall, the average hallucination rate runs roughly between  across models. %and $6-15\%$ across languages. 
%         
%, we show the averages and the standard deviations for our estimations per model and language. 
%
% Gemma 
%Scale Languages, Hallucinate More
For each LLM, we have 15 estimates (3 HD model instances $\times$ 5 generations by the LLM) of $\mathit{HR}$ (averaged across all languages): we apply the Student's t-test to determine if the differences between models (smaller and larger) significantly differ. Figure \ref{fig:paired-bar-graph} summarizes the results. The difference between the two EuroLLM variants is not significant; larger Gemma model hallucinates more (significantly), but the HR are low for both variants; for Llama and Qwen, the smaller models hallucinate significantly more. Finally, we aggregate the estimates across all ``small'' models (1.7-3B) and all ``large'' models (7-9B) and see that, overall (column ``Overall'' in Figure \ref{fig:paired-bar-graph}), smaller LLMs hallucinate significantly more ($p = 0.01$). This agrees with \newcite{wei2024long} who report larger models to be more truthful in long-form answer generation. 

%However, this behavior is not consistent in EuroLLM and more prominently in Gemma where the opposite seems to be the case which shows that, a larger model may have a higher knowledge capacity but if the ground truth function $f$ is not learned \citep{xu2024hallucination}, changing the size of the model is futile and scaling up can make the model more unreliable \citep{chang2024how, mahapatra2024impact}. 

\rparagraph{More Languages, More Hallucination?} Figure \ref{fig:rate_vs_supported_language} compares LLMs' hallucination rates against their declared number of supported languages. here we a surprising trend see that LLMs that support more languages tend to hallucinate more (e.g., EuroLLM supports 35 languages, whereas Gemma is declared to support English only)---the correlation is strong and significant ($r=0.88, p=0.049$).

\rparagraph{Say Less, Hallucinate Less?} Intuitively, one would expect LLMs' hallucination rates to be larger for languages in which they are less competent (i.e., seen the least in pretraining and instruction-tuning). 
%would hallucinate most in the languages that they have seen the least during their training. 
Surprisingly, however, we do not find this to be the case for any of the models. E.g., we observe the lowest hallucination rate for Sindhi ($5.83\%$ of tokens are hallucinated), a language with merely 18,000 Wikipedia articles and largest hallucination rate for Hebrew ($16.81\%$). Across all 30 languages, however, we find \textit{no} correlation between the hallucination rates and measures of language ``resourceness'': (i) proportion of language-specific data in Common Crawl and (ii) number of articles in the language-specific Wikipedia. 
As illustrated in Figure \ref{fig:overall}, we do observe that LLMs generate longer responses for languages in which they are more competent---this entails a larger number of hallucinated tokens for longer responses, but not (necessarily) a larger (per-token) hallucination rate (recall that we account for the response length in Eq.\,\ref{eq:HR_adjusted}). Indeed, we observe no correlation whatsoever between the response length and hallucination \textit{rates} across languages ($r = -0.05$). This suggests that a trade-off between the answer length and the amount (not rate!) of hallucinations is a largely language-independent property of LLMs.

\begin{figure*}[t]
    \centering
    \includegraphics[width=\textwidth]{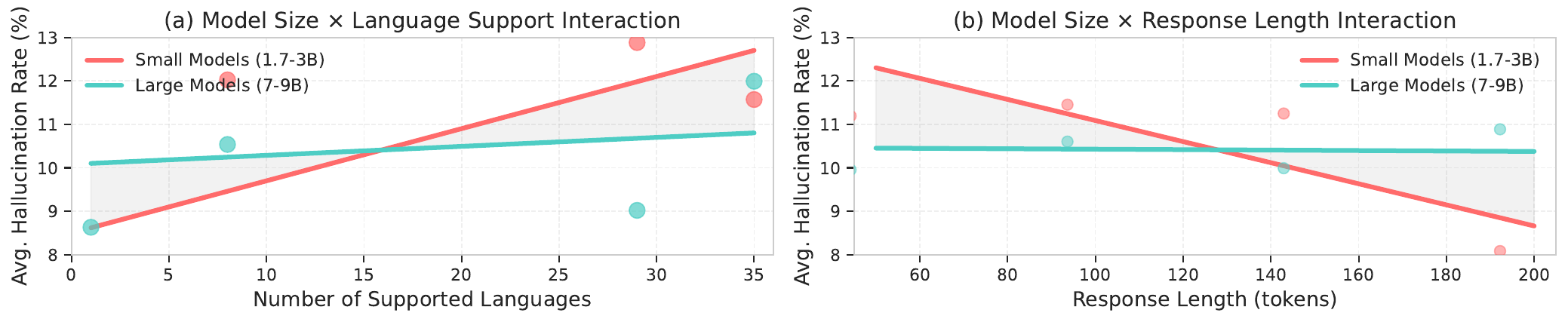}
    \caption{Significant interaction effects from the linear mixed effects model. (a) Model Size $\times$ Language Support interaction ($\beta = 1.33$, $p < 0.001$). (b) Model Size $\times$ Response Length interaction ($\beta = -1.02$, $p = 0.001$). Dots represent observed means; shaded areas show the magnitude of difference between model sizes.} 
    \label{fig:mixed-effects-interactions}
    \vspace{-0.5em}
\end{figure*}

\subsection{Mixed Effect Modeling} 

To understand the potential dependencies between factors identified in \S\ref{sec:final-estimates}, we conducted a linear mixed effects analysis with hallucination rate as the dependent variable. The model included fixed effects for model size (small: 1.7-3B vs. large: 7-9B parameters), number of supported languages, and response length, along with all two-way interactions. The interaction model significantly improved fit over the main effects model ($\text{Likelihood Ratio} = 22.14$, $p < 0.001$), showing that these factors do not operate independently. 

Figure~\ref{fig:mixed-effects-interactions} reveals two significant interactions: (1) \textbf{Model Size $\times$ Language Support} ($\beta = 1.33$, $p < 0.001$): the hallucination gap between small and large models widens as language support increases---small models struggle disproportionately with multilingual capabilities; and (2) \textbf{Model Size $\times$ Response Length} ($\beta = -1.02$, $p = 0.001$): larger models excel particularly for longer responses, while this advantage diminishes for shorter outputs. These findings reveal that model size acts as a critical moderator---larger models not only hallucinate less overall but are also more robust to the challenges posed by multilingual support and longer generation tasks, suggesting that scaling model parameters provides compound benefits beyond simple performance improvements.

\section{Conclusion}

%Little is known about the amount of hallucinations multilingual LLMs produce ``in the wild''. 
We presented the first effort towards understanding how much multilingual LLMs hallucinate in a realistic setting. To this end, we proposed a novel framework for hallucination rate estimation, which adjusts the number of detected hallucinations based on the detector's performance resulting in more reliable rate estimates. We trained a series of multilingual detection models, and measured their precision and recall scores on our newly created \textsc{mFAVA} datasets across 30 languages. To estimate hallucinations, we build a novel synthetic open-domain knowledge-intensive QA dataset for which we collected answers from eleven open-source LLMs. Our findings indicate that smaller models and models that cover more languages hallucinate significantly more, and that model response-length does not correlate with hallucination rate. Lastly, linear mixed effects analysis reveals that these factors do not operate independently---model size acts as a critical moderator, with smaller models suffering disproportionately when supporting multiple languages and generating longer responses.

\section*{Limitations}
\label{sec:lim}
%We acknowledge the following limitations of our work: \textcolor{red}{(1) 
We acknowledge that our method of using GPT-4 to insert synthetic hallucinations may not perfectly replicate natural model errors. However, this approach was chosen due to the immense difficulty and expense of manually curating such a dataset in 30 languages (detailed in §\ref{sec:mfava}). Crucially, our findings in Table \ref{tab:fool-rating} indicate that more than 50\% of these synthetic hallucinations were still perceived as convincing and realistic.

We adopted the common translation-train approach and thus used MT to translate the original \textsc{FAVA} into our 30 target languages. While one may argue that we thus add some noise to the training process resulting in unreliable detectors, recall that we are not opting for the highest possible detection performances, but rather interested in obtaining reliable performance estimates. 

We only have gold annotations for 5 languages. Here, one might argue that, thus, our performance estimates might be unreliable. This is why in \S\ref{subsec:silver-vs-gold}, we compare estimates obtained on \textsc{mFAVA}-Silver with ones obtained on \textsc{mFAVA}-Gold and show that silver annotations can serve as a reliable proxy. 

For our hallucination evaluation, we only manually check a subset of the Arabic, Chinese, German, Russian, and Turkish queries to ensure that the answers to the synthetic prompts are present in the Wikipedia references. The high rate of 98\% we observed makes us confident that the potential error we introduce via such ``non-grounded'' questions for other languages is negligible, especially for high-resource languages. We still acknowledge, however, that the Wikipedia articles we use might be limited in terms of the knowledge they cover~\citep{kim-etal-2024-analysis}, this is why we carefully filter via minimum length and collaborative Wikipedia depth towards higher-quality articles with high coverage.

Finally, we deliberately limited the scope of this work to assessing factual correctness and we do not cover factual coverage. We decided to do so as quantifying hallucinations in long-form generation is already difficult for English \citet{xu2023critical, min2023factscore, wei2024long} and more so in non-english languages \citep{kim-etal-2024-analysis}, and currently, resources for assessing multilingual factual coverage are still lacking.

\section*{Acknowledgments}

This work was supported by the Alcatel-Lucent
Stiftung and Deutsches Stiftungszentrum through
the grant “Equitably Fair and Trustworthy
Language Technology” (EQUIFAIR, Grant Nr.
T0067/43110/23). The work of Anne Lauscher  is funded under the Excellence Strategy of the German Federal Government and States. The authors gratefully acknowledge the computing time granted by the John von Neumann Institute for Computing (NIC) and provided on the supercomputer JURECA \citep{JURECA} at Jülich Supercomputing Centre (JSC).

\bibliography{custom}

\appendix

\section{Appendix}\label{sec:appendix}

\subsection{Choice of languages}\label{appendix:choice-languages}
Initially, we wanted to cover all 14 language families based on Glottolog 5.0 \citep{nordhoff-hammarstrom-2012-glottolog}), however, as we progressed through the languages, we found that even the best closed-source LLMs like GPT-4 and Gemini are bad at generating text in low-resource languages (e.g. Amharic, Aymara, Hausa and Tamil) and we could not employ LLMs to generate and annotate a silver hallucination detection dataset in these languages. See Table \ref{tab:language_classification} for 30 languages.

\subsection{Annotation Process}\label{appendix:annotation-process}

We provide the FAVA seed passage generations and hallucination insertion prompts in Tables \ref{tab:prompt-ent}, \ref{tab:prompt-rela}, \ref{tab:prompt-contra}, \ref{tab:prompt-subj}, \ref{tab:prompt-unv}, \ref{tab:prompt-inv}.

\begin{figure*}[h]
    \centering
    \includegraphics[width=\textwidth]{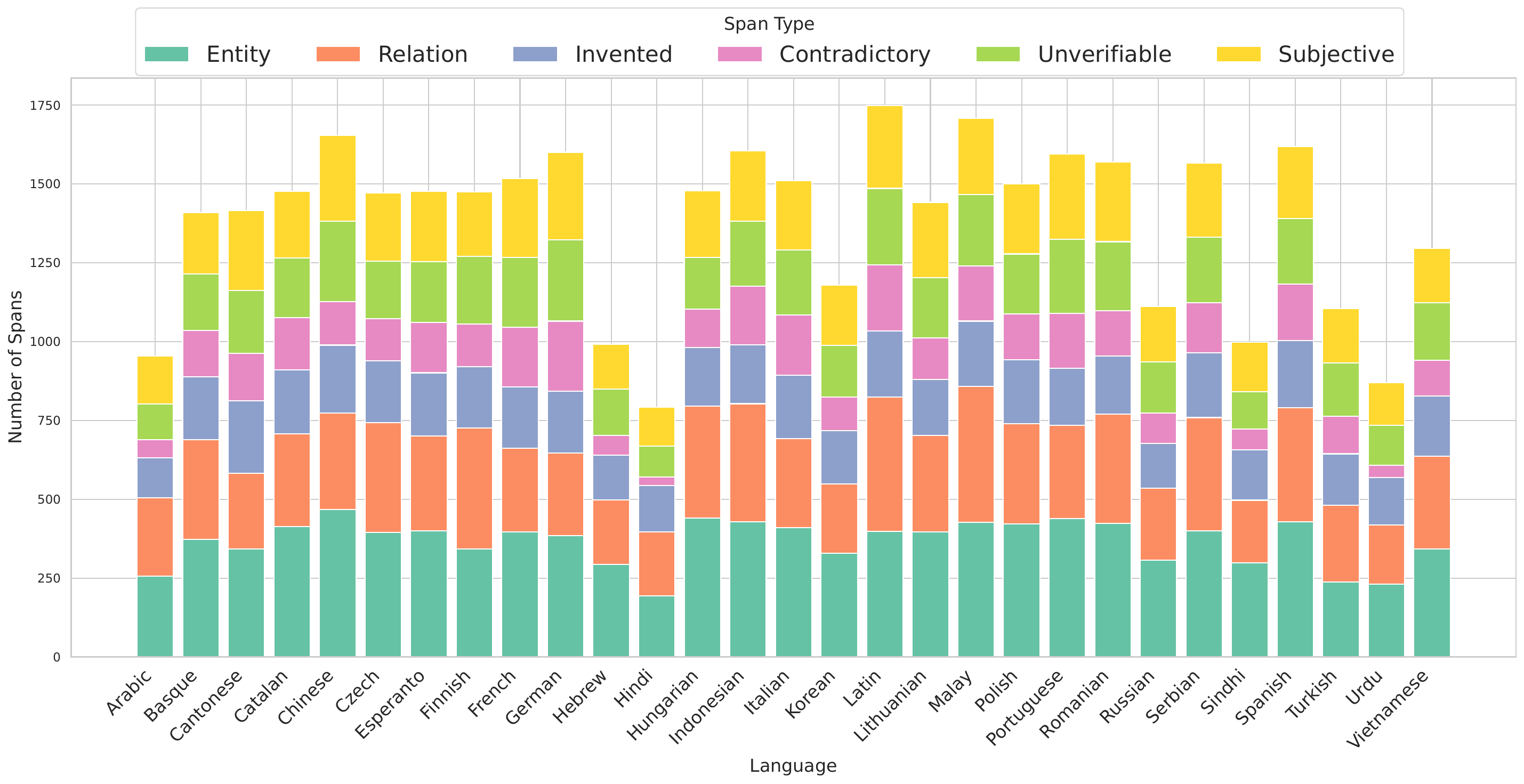}
    \caption{Distribution of 6 labels across 30 languages in \textsc{mFava-Silver} dataset.}
    \label{fig:silver-stacked-bar-chart}
\end{figure*}

\paragraph{Cost of Silver Annotations} The total cost for generating silver data for 30 languages using GPT-4 was $\sim$\$2,310 with $\sim$\$77 per language. Distribution of categories across 30 languages is provided in Table \ref{tab:category_percentages} and and per language label distribution is provided in Figure \ref{fig:silver-stacked-bar-chart}.

\begin{table}[h]
    \centering
    \resizebox{\columnwidth}{!}{%
    \begin{tabular}{lcccccc}
    \toprule
     & \textbf{ENT} & \textbf{REL} & \textbf{INV} & \textbf{CON} & \textbf{UNV} & \textbf{SUB}  \\
    \midrule
    \textbf{Count} & 11143 & 9036 & 5649 & 4024 & 5670 & 6396  \\
    \bottomrule
    \end{tabular}%
    }
    \caption{Distribution of categories across 30 languages in silver set.}
    \label{tab:category_percentages}
\end{table}

\paragraph{Gold Annotations:} The annotators were sourced through prolific platform. Each annotator was screened on 10 samples and if they met the threshold of 40\% agreement with the silver annotation, they were invited to participate in the full study. 

It is worth noting that as the hallucination annotation task for longform QA is very cognitively demanding, it took us a long time to find annotators who could do the task correctly with high-effort. Most of the time, annotators who passed the screening test, decided to leave the study because of the high effort requirement of the task even though our study was paying above minimum wage (14 \$/hr) for the full study. Moreover, Table \ref{fig:agreement-metrics} reveals that Inter-Annotator Agreement (IAA) and Silver-Gold agreement for category annotations are both below 80\%, underscoring the inherent difficulty of this task. This challenge is further reflected in our token-level agreement scores, which are impacted by minor inconsistencies in annotator decisions regarding minimal span selection.

\begin{figure*}[h]
    \centering
    % First subfigure
    \begin{subfigure}{\textwidth}
         \centering
         \includegraphics[scale=0.8]{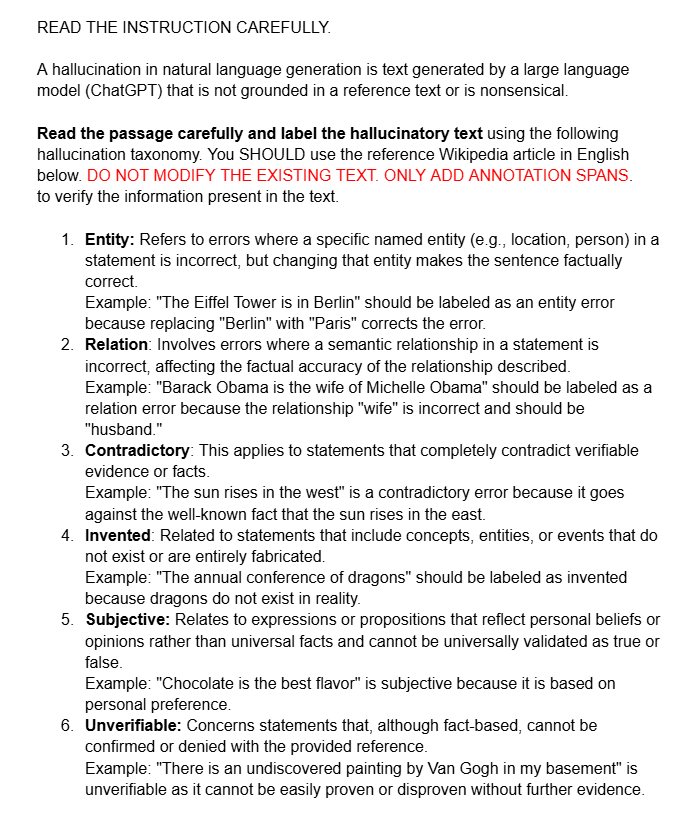}
         \label{fig:annotation-instruction-1}
    \end{subfigure}
    % Add some vertical space between subfigures (optional)
    \vspace{1em}
    
    % Second subfigure
    \begin{subfigure}{\textwidth}
         \centering
         \includegraphics[scale=0.8]{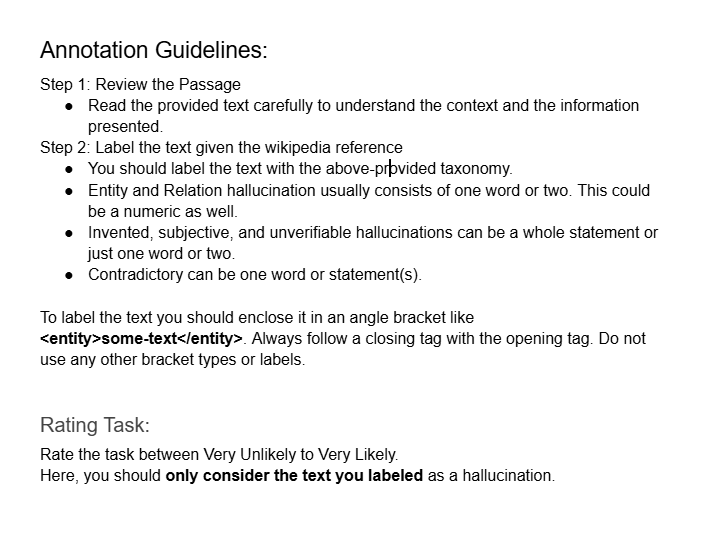}
         \label{fig:annotation-instruction-2}
    \end{subfigure}
    
    \vspace{1em}
    \caption{Annotation Instructions.}
    \label{fig:annotation-instruction}
\end{figure*}

The total cost of the gold annotations was (including platform and annotation fees) \textbf{\$4581} where each annotator was paid 14 \$/hr. All the annotators were at least bachelor's level and bilingual because in addition to understanding their own language (e.g. Arabic) they also needed to understand the task instructions and Wikipedia content (in English). The task instructions are given in Figure \ref{fig:annotation-instruction}. Each annotator was asked if they consent to storing their prolific IDs during manual and automatic assessment stage. Following the assessment, their prolific IDs were deleted.

We will release \textsc{mFava} data under an open scientific licensing.

\subsection{Training Details}\label{appendix:training-details}

All the classifiers were trained utilizing the Bi-LLM \citep{li2023label} and transformers \citep{wolf2019huggingface} library. The models were trained with three seeds (42, 47, 49) on 4xH100 until convergence. Seeds are set for \textit{torch.manual\_seed()} and \textit{random.seed()}. The exact hyper-parameters are given in the Table \ref{tab:training_details}. Total GPU hours:  1134. 

\begin{table}[h]
    \centering
    \small  % Reduce font size to fit single column
    \begin{tabular}{@{}ll@{}}
        \toprule
        \textbf{Parameter} & \textbf{Value} \\
        \midrule
        Translate Train-Val Split & 70:30 \\
        Seeds & [42, 47, 49] \\
        Quantization & 4-bit BF16 \\
        Model & Llama-3-8B (base) \\
        GPUs & 4$\times$ H100 \\
        LoRA $r$ & 32 \\
        LoRA $\alpha$ & 32 \\
        LoRA Dropout & 0.05 \\
        LoRA Target Modules & All \\
        Epochs & $\sim$2 (until convergence) \\
        Input Length & 4096 \\
        Learning Rate & $1 \times 10^{-4}$ \\
        Weight Decay & 0.01 \\
        Batch Size & 8 \\
        Gradient Accumulation & 8 \\ 
        \bottomrule
    \end{tabular}
    \caption{Training Details}
    \label{tab:training_details}
\end{table}

\subsection{Adjusting for $\mathit{P_l}$ and $\mathit{R_l}$}\label{appendix:derivation}
The hallucination rate  \(\mathit{HR}_{est,l}\) for a given language \(\mathit{l}\), is defined as the ratio of hallucinated tokens detected by the model (\(\mathit{H}_{\text{detected}, l}\)) to the total number of generated tokens (\(\mathit{N}_l\)):

\begin{small}
\begin{equation}\label{eq:HR_l}
\mathit{HR}_{l} = \frac{\mathit{H}_{\text{det}, l}}{\mathit{N}_l}\,.
\end{equation}
\end{small}

To refine this rate, we adjust for the detection model's precision (\(\mathit{P}_l\)) and recall (\(\mathit{R}_l\)). Precision is defined as:

\begin{small}
\begin{equation}\label{eq:precision}
\mathit{P}_l = \frac{\mathit{TP}_l}{\mathit{TP}_l + \mathit{FP}_l}
\end{equation}
\end{small}

where (\(\mathit{TP}_l\)) \(\mathit{FP}_l\) denote true and false positives respectively. Rearranging this equation gives the number of true positives:

\begin{small}
\begin{equation}\label{eq:TP_l}
\mathit{TP}_l = \mathit{P}_l \cdot \mathit{HR}_{\text{det}, l}
\end{equation}    
\end{small}

Recall is defined as:

\begin{small}
\begin{equation}\label{eq:recall}
\mathit{R}_l = \frac{\mathit{TP}_l}{\mathit{TP}_l + \mathit{FN}_l}
\end{equation}    
\end{small}

where \(\mathit{FN}_l\) denotes false negatives. The total number of corrected hallucinations (\(\mathit{HR}_{\text{est}, l}\)) can thus be expressed as:

\begin{small}
\begin{equation}\label{eq:H_true}
\mathit{HR}_{\text{est}, l} = \mathit{TP}_l + \mathit{FN}_l = \frac{\mathit{TP}_l}{\mathit{R}_l}
\end{equation}
\end{small}

Substituting Equations~\ref{eq:TP_l} in ~\ref{eq:H_true}, we derive the $\mathit{H}_{\text{est}, l}$ as:

\begin{small}
\label{equation:adjusted-formula}
\begin{equation}\label{eq:HR_adjusted_app}
\mathit{H}_{\text{est}, l} = \frac{\mathit{P}_l \cdot \mathit{H}_{\text{det}, l}}{\mathit{R}_l}
\end{equation}    
\end{small}

By incorporating the model's $P_{l}$ and $R_{l}$, our estimation framework effectively corrects for the imperfections of a hallucination detector. When estimating the hallucination rate $\mathit{HR}_{l}$ on a large corpus (see \S\ref{sec:estimation_methodology}), EQ \ref{eq:HR_adjusted_app} provides a reliable measure of the true number of hallucinations. This accounts for the detector erroneously flagging $1-P_{l}\%$ of its identified instances and failing to capture $1-R_{l}\%$ of genuine hallucinations.

\subsection{Hallucination Evaluation Dataset}\label{appendix:evaluation-dataset}
To construct the hallucination evaluation dataset, we aimed to scrap $\sim$ 1000 articles per language with more than 2000 characters. However, problem with non-English languages (especially moderate-low resource) is that  $\geq2000$ character articles can be scarce. Furthermore, sometimes Wikipedia has articles tagged as \textit{Unreferenced, Failed Verification,} or \textit{Under Construction} which flag the article as unfinished or not factually verified. Such tags are very prominent in languages other than English and we do not include such articles in our dataset.

We use Wikipedia article summary (text before the first heading) as references and prompt gpt-4 to generate 2 knowledge-intensive queries per article. Sometimes, it generated only one query even though we explicitly state to generate two queries. We did not prompt the GPT-4 again to generate the second query due to budget constraints. The total cost to generate prompts for 31 languages is \$192. Per language statistics can be found in Table \ref{tab:hallucination-eval-dataset}. We release hallucination evaluation data under an open scientific licensing.

\begin{table}[ht]
\centering
\fbox{%
  \begin{minipage}{0.8\columnwidth}
    Given the following reference in language \textit{<language name>}:
    
    Generate \textbf{two knowledge-intensive queries} in \texttt{<language name>}. Ensure the questions are concise but knowledge-intensive. The questions should require thorough reading of the reference text to answer. Separate the questions with a newline.
  \end{minipage}
}
\caption{Prompt for generating knowledge-intensive queries.}
\label{tab:prompt}
\end{table}

\subsection{Response Collection for Hallucination Evaluation Dataset} \label{appendix:response-collection}

We collect LLM responses on 5 seeds: 42, 43, 44, 47, 49 for 6 LLM model\footnote{We comply with licensing agreement for each of the LLMs we use.}. The generation configurations that we used are provided in the \textit{generation\_config.json} in model repositories on huggingface. Seeds are set for \textit{torch.manual\_seed()} and \textit{random.seed()}. 

\subsection{Manual Analysis}\label{appendix:manual-analysis}
To further check the quality and informativeness (See \S\ref{appendix:infomativeness} for definitions of informativeness) of the responses, we manually analyze 60 responses from  Aya-23-8B for German and Arabic, two languages for which we have gold annotations. Overall, 10\% of the responses had repetitive words and sentences and 5\% of the responses were \textit{I don't know} responses. For the remainder of the samples, the responses were fluent and long and were relevant to the input prompt. 

\subsection{Informativeness} \label{appendix:infomativeness}

Currently, there is no agreed-upon definition of informativeness. \citet{lin2022truthfulqa} considers a response to be informative if it is potentially relevant to the question and \citet{wei2024long} considers a response to be informative if it has a certain number of supporting facts from the reference text.

\begin{table*}[h]
\centering
\small
\begin{tabular}{lccc}
\toprule
\textbf{Language} & \textbf{Language Family} & \textbf{Script} & \textbf{Test-Set} \\
\midrule
% Gold test sets at the top, sorted alphabetically
Arabic        & Afro-Asiatic (Semitic)      & Arabic        & Gold    \\
Chinese       & Sino-Tibetan (Sinitic)      & Chinese (Han) & Gold    \\
German        & Indo-European (Germanic)    & Latin         & Gold    \\
Russian       & Indo-European (Slavic)      & Cyrillic      & Gold    \\
Turkish       & Turkic (Common Turkic)      & Latin         & Gold    \\
\cmidrule{1-4} % Partial horizontal line separating Gold and Silver test sets

% Silver test sets, sorted alphabetically
Basque        & Language Isolate             & Latin         & Silver  \\
Cantonese     & Sino-Tibetan (Sinitic)       & Chinese (Han) & Silver  \\
Catalan       & Indo-European (Romance)      & Latin         & Silver  \\
Czech         & Indo-European (Slavic)       & Latin         & Silver  \\
Esperanto     & Constructed                  & Latin         & Silver  \\
Finnish       & Uralic (Finnic)              & Latin         & Silver  \\
French        & Indo-European (Romance)      & Latin         & Silver  \\
Hebrew        & Afro-Asiatic (Semitic)       & Hebrew        & Silver  \\
Hindi         & Indo-Aryan                   & Devanagari    & Silver  \\
Hungarian     & Uralic (Ugric)               & Latin         & Silver  \\
Indonesian    & Austronesian (Malayo-Polynesian) & Latin     & Silver  \\
Italian       & Indo-European (Romance)      & Latin         & Silver  \\
Japanese      & Japonic                      & Kanji         & Silver  \\
Korean        & Koreanic                     & Hangul        & Silver  \\
Latin         & Indo-European (Italic)       & Latin         & Silver  \\
Lithuanian    & Indo-European (Slavic)       & Latin         & Silver  \\
Malay         & Austronesian (Malayo-Polynesian) & Latin     & Silver  \\
Polish        & Indo-European (Slavic)       & Latin         & Silver  \\
Portuguese    & Indo-European (Romance)      & Latin         & Silver  \\
Romanian      & Indo-European (Romance)      & Latin         & Silver  \\
Serbian       & Indo-European (Slavic)       & Cyrillic      & Silver  \\
Sindhi        & Indo-Aryan                   & Arabic        & Silver  \\
Spanish       & Indo-European (Romance)      & Latin         & Silver  \\
Urdu          & Indo-Aryan                   & Arabic        & Silver  \\
Vietnamese    & Austroasiatic (Vietic)       & Latin         & Silver  \\
\bottomrule
\end{tabular}
\caption{Classification of languages by language family (based on Glottolog 5.0), script, and test-set status. Gold test sets are available for 5 languages, while the rest have silver test sets.}
\label{tab:language_classification}
\end{table*}

\begin{table*}[h]
    \centering
    \footnotesize
    \renewcommand{\arraystretch}{1.1}
    \begin{tabular}{lcccccc}
        \toprule
        \textbf{Model} & \textbf{max\_new\_tokens} & \textbf{temperature} & \textbf{top\_p} & \textbf{top\_k} & \textbf{repetition\_penalty} & \textbf{do\_sample} \\
        \midrule
        Llama-3.x   & 1024  & 0.6   & 0.9  & --  & -- & True \\
        Aya     & 1024 & --   & 0.3 & --  & -- & True \\
        Qwen-2.5  & 1024  & 0.7   & 0.9  & 20  & 1.05 & True \\
        Mistral & 1024 & --   & -- & 50  & --   & True \\
        Gemma-2   & 1024  & -- & -- & -- & -- & True \\
        EuroLLM    & 1024  & -- & -- & --  & --  & True \\
        \bottomrule
    \end{tabular}
    \caption{Huggingface \textsc{model.generate()} parameters for each model family. -- indicate default is used. Generation configurations are provided in model's respective HuggingFace \citep{wolf2019huggingface} repositories}
    \label{tab:generation-config}
\end{table*}

\begin{table*}[ht]
\centering
\begin{tabular}{lrrr}
\hline
\textbf{Language} & \textbf{Unique Categories} & \textbf{Total Articles} & \textbf{Total Queries} \\
\hline
Arabic      & 537  & 959  & 1907 \\
Basque      & 486  & 938  & 1872 \\
Cantonese   & 261  & 401  & 793  \\
Catalan     & 359  & 989  & 1976 \\
Chinese     & 712  & 977  & 1939 \\
Czech       & 720  & 988  & 1975 \\
Esperanto   & 608  & 956  & 1912 \\
French      & 332  & 987  & 1973 \\
Finnish     & 549  & 995  & 1972 \\
German      & 797  & 984  & 1967 \\
Hebrew      & 660  & 999  & 1991 \\
Hindi       & 153  & 186  & 367  \\
Hungarian   & 745  & 992  & 1964 \\
Indonesian  & 457  & 958  & 1913 \\
Italian     & 678  & 988  & 1974 \\
Japanese    & 667  & 999  & 1991 \\
Korean      & 539  & 747  & 1488 \\
Latin       & 334  & 465  & 916  \\
Lithuanian  & 711  & 946  & 1888 \\
Malay       & 442  & 778  & 1556 \\
Polish      & 889  & 1000 & 1998 \\
Portuguese  & 390  & 955  & 1909 \\
Romanian    & 351  & 811  & 1618 \\
Russian     & 462  & 999  & 1996 \\
Spanish     & 938  & 977  & 1952 \\
Serbian     & 386  & 798  & 1587 \\
Sindhi      & 224  & 519  & 1029 \\
Turkish     & 660  & 856  & 1650 \\
Urdu        & 567  & 878  & 1749 \\
Vietnamese  & 326  & 660  & 1311 \\
\hline
\textbf{Total} & \textbf{15,940} & \textbf{25,685} & \textbf{51,133} \\
\hline
\end{tabular}
\caption{Per language statistics for hallucination evaluation dataset.}
\label{tab:hallucination-eval-dataset}
\end{table*}

\begin{table*}[h]
    \centering
    \begin{tabular}{lccc}
        \toprule
        \textbf{Language} & \textbf{Precision (\%)} & \textbf{Recall (\%)} & \textbf{F1 Score (\%)} \\
        \midrule
        \multicolumn{4}{c}{\textbf{GOLD}} \\
        \midrule
        Arabic (Gold)    & 73.98 & 53.40 & 61.63 \\
        Chinese (Gold)   & 70.73 & 53.93 & 58.79 \\
        German (Gold)    & 58.19 & 74.06 & 65.05 \\
        Turkish (Gold)   & 79.67 & 66.95 & 72.57 \\
        Russian (Gold)   & 63.18 & 68.46 & 65.53 \\
        \hline
        Average          & 69.15 & 63.36 & 64.71 \\
        \midrule
        \multicolumn{4}{c}{\textbf{SILVER}} \\
        \midrule
        Arabic           & 93.28 & 74.81 & 82.59 \\
        Chinese          & 80.33 & 66.28 & 69.77 \\
        German           & 91.64 & 87.77 & 89.50 \\
        Turkish          & 89.58 & 83.92 & 86.43 \\
        Russian          & 93.05 & 86.04 & 89.15 \\
        Basque           & 87.22 & 74.46 & 79.80 \\
        Cantonese        & 78.49 & 49.40 & 56.12 \\
        Catalan          & 94.70 & 87.46 & 90.85 \\
        Czech            & 93.99 & 84.75 & 89.00 \\
        Esperanto        & 94.28 & 86.53 & 90.05 \\
        French           & 91.58 & 89.37 & 90.31 \\
        Finnish          & 86.67 & 84.26 & 85.15 \\
        Hebrew           & 82.75 & 32.97 & 44.19 \\
        Hindi            & 68.01 & 68.48 & 66.77 \\
        Hungarian        & 92.35 & 74.29 & 81.93 \\
        Indonesian       & 92.12 & 85.75 & 88.72 \\
        Italian          & 93.76 & 87.26 & 90.28 \\
        Korean           & 86.39 & 79.11 & 82.31 \\
        Japanese         & 77.06 & 61.03 & 67.15 \\
        Lithuanian       & 90.48 & 75.39 & 81.81 \\
        Malay            & 86.15 & 68.96 & 75.73 \\
        Portuguese       & 95.80 & 86.77 & 90.94 \\
        Serbian          & 86.16 & 76.75 & 79.91 \\
        Sindhi           & 82.00 & 69.38 & 74.36 \\
        Spanish          & 95.86 & 85.34 & 90.14 \\
        Vietnamese       & 89.35 & 84.57 & 86.71 \\
        Urdu             & 88.82 & 72.32 & 79.39 \\
        \hline
        Average          & 88.22 & 76.42 & 80.71 \\
        \bottomrule
    \end{tabular}
    \caption{Precision, Recall, and F1 scores for all languages, including GOLD scores for five languages.}
    \label{tab:hallucination-metrics-extended}
\end{table*}

\begin{figure*}[h]
    \centering
    % First subfigure
    \begin{subfigure}{\textwidth}
         \centering
         \includegraphics[width=\textwidth]{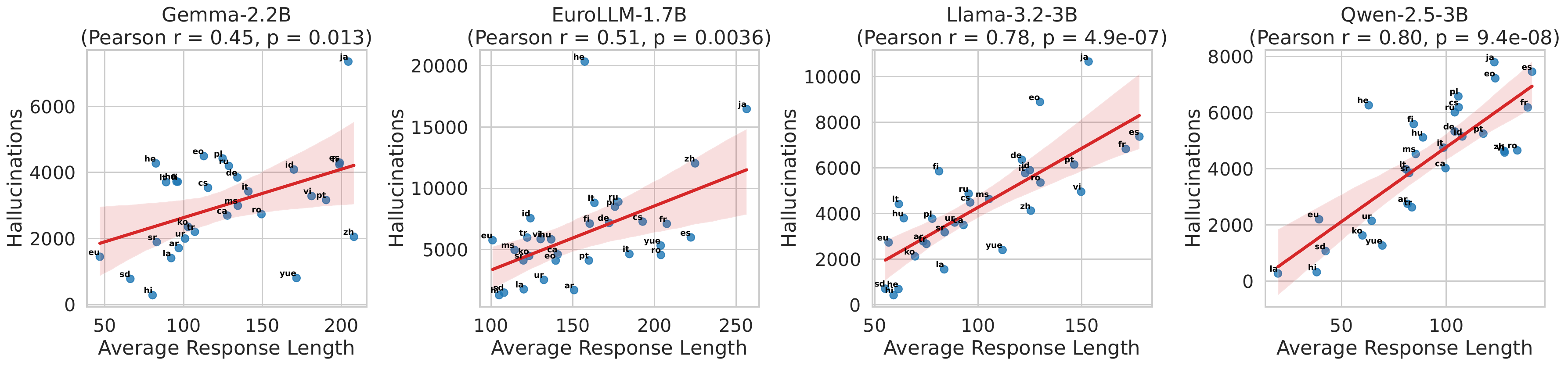}
         \caption{Hallucinations vs response length correlation of smaller models.}
         \label{fig:small-correlation-map-1}
    \end{subfigure}
    
    % Add some vertical space between subfigures (optional)
    \vspace{1em}
    
    % Second subfigure
    \begin{subfigure}{\textwidth}
         \centering
         \includegraphics[width=\textwidth]{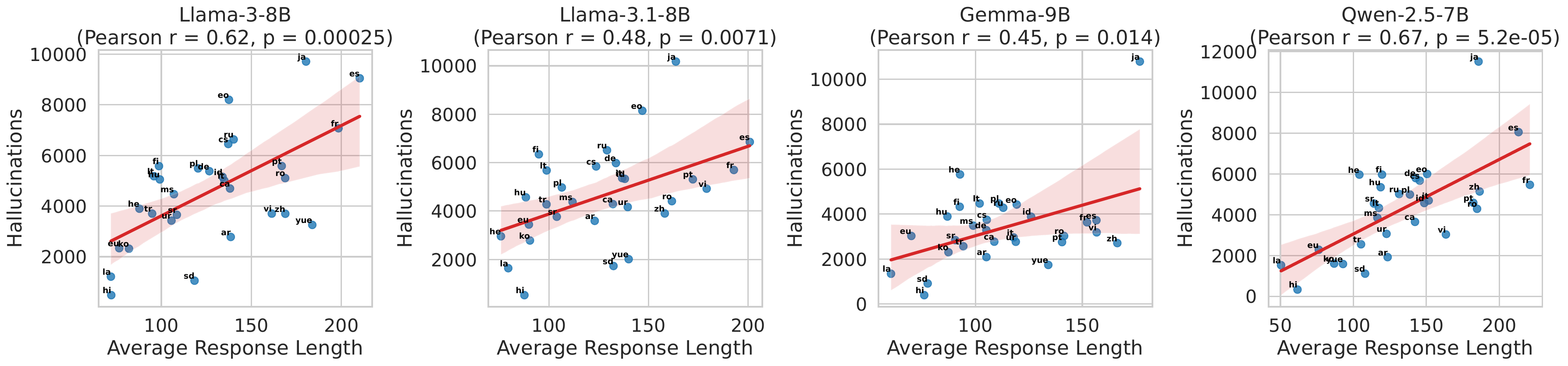}
         \caption{Hallucinations vs response length correlation of bigger models.}
         \label{fig:big-correlation-map-2}
    \end{subfigure}
    
    \vspace{1em}
    
    % Third subfigure
    \begin{subfigure}{\textwidth}
         \centering
         \includegraphics[width=\textwidth]{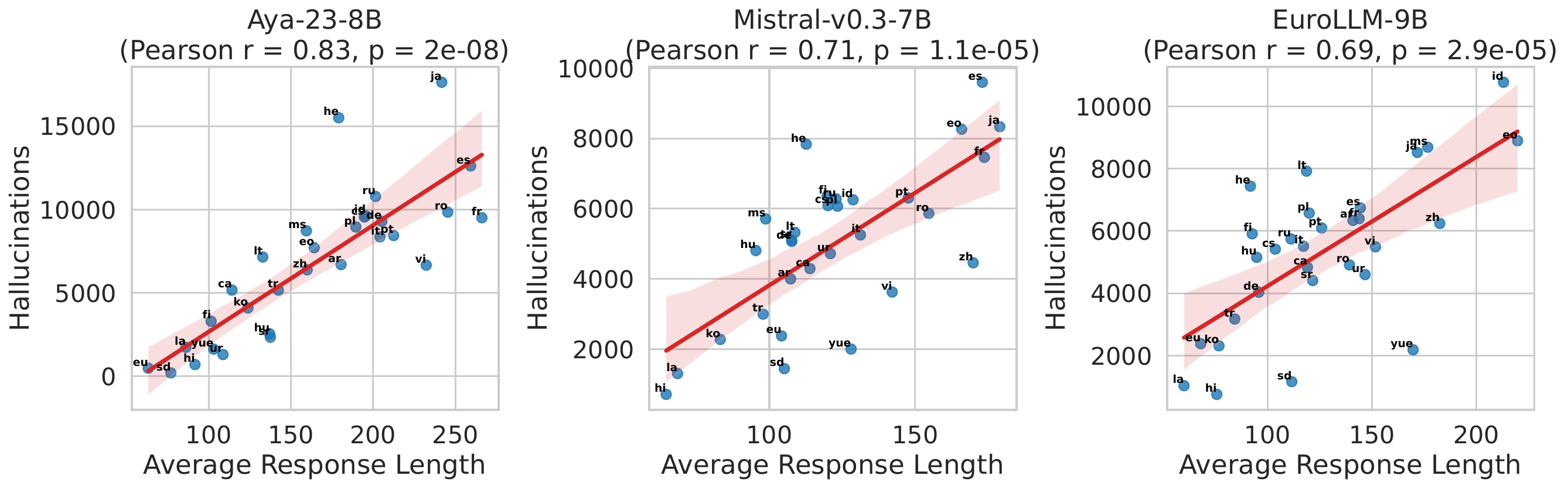}
         \caption{Hallucinations vs response length correlation of bigger models.}
         \label{fig:big-correlation-map-3}
    \end{subfigure}
    \caption{Per model correlations between hallucinations and response length.}
    \label{fig:stacked-correlation}
\end{figure*}

\onecolumn
\appendix % Or just start a new section, e.g., \section{Prompts}
           % Using \appendix is common for supplementary material like full prompts.

\begin{table}[b]
\centering
\fbox{%
  \begin{minipage}{\textwidth}
    For the following instruction, generate the whole output in \{lang\} language even though the instruction and input is in English.\newline
    Given a passage answer the question in 3-5 sentences using only the information presented in the reference passage.\newline
    Question: [QUESTION\_TEXT]\newline
    Reference Passage: [REFERENCE\_CONTENT]
  \end{minipage}
}
\caption{Prompt for seed passage generation.}
\label{tab:prompt}
\end{table}

\begin{table}[b]
\centering
\fbox{%
  \begin{minipage}{\textwidth}
  \small
    For the following instruction, generate the whole output in \{lang\} language even though the instruction and input is in English.\newline
    Given a passage with possibly already inserted error tokens wrapped in \textless{}relation\textgreater{}, \textless{}contradictory\textgreater{}, \textless{}unverifiable\textgreater{}, \textless{}subjective\textgreater{}, or \textless{}invented\textgreater{}, insert entity errors in the passage below, wrapped in tokens to make the passage factually incorrect. Ensure these insertions are outside these existing \textless{}\textgreater{} tags, and don't modify the \textless{}\textgreater{} tags at all. The error is defined as such:\newline
    1. entity errors (\textless{}entity\textgreater{}): a small part of a sentence, often an entity (e.g., location name), is incorrect (usually 1-3 words). Entity errors often involve noun phrases or nouns.\newline
    Example 1: Messi is an \textless{}entity\textgreater{}\textless{}/entity\textgreater{} soccer player.\newline
    Example 2: Selena Gomez was born on \textless{}entity\textgreater{}\textless{}/entity\textgreater{} 22.\newline
    Example 3: India's population is \textless{}entity\textgreater{}\textless{}/entity\textgreater{} billion people.\newline
    Now, insert entity error tokens in the given passage but make sure that you don't modify anything inside any already existing \textless{}\textgreater{} error tokens, only add entity errors with \textless{}entity\textgreater{}\textless{}/entity\textgreater{} tokens outside the already existing \textless{}\textgreater{} tags. \newline
    \#\#\newline
    Paragraph: [PASSAGE\_CONTENT]\newline
    Edited:
  \end{minipage}
}
\caption{FAVA Prompt for entity hallucination insertion.}
\label{tab:prompt-ent}
\end{table}

\begin{table}[b]
\centering
\fbox{%
  \begin{minipage}{\textwidth}
  \small
    For the following instruction, generate the whole output in \{lang\} language even though the instruction and input is in English.\newline
    Given a passage with possibly already inserted error tokens wrapped in \textless{}entity\textgreater{}, \textless{}contradictory\textgreater{}, \textless{}unverifiable\textgreater{}, \textless{}subjective\textgreater{}, or \textless{}invented\textgreater{}, insert relation errors, outside the already inserted tokens without modifying the content within already existing tokens. Wrap the relational errors in tokens to make the passage factually incorrect. The error is defined as such:\newline
    1. relational error (\textless{}relation\textgreater{}): a sentence is partially incorrect as a small part (usually 1 - 3 words). Relational errors often involve verbs and are often the opposite of what it should be.\newline
    Example 1: FDA \textless{}relation\textgreater{}\textless{}/relation\textgreater{} pfizer COVID-19 Vaccine.\newline
    Example 2: Rishi Sunak \textless{}relation\textgreater{}\textless{}/relation\textgreater{} his role as Prime Minister in 2022.\newline
    Example 3: Millie Bobbie Brown has also starred in several popular movies, including “Godzilla vs. Kong” and “Enola Holmes” which she also \textless{}relation\textgreater{}\textless{}/relation\textgreater{}.\newline
    Now, insert relation error tokens in the given passage but make sure that you don't modify anything inside any already existing \textless{}\textgreater{} error tokens, only add relational errors with \textless{}relation\textgreater{}\textless{}/relation\textgreater{} tokens outside the already existing \textless{}\textgreater{} tags. 
    \#\#\newline
    Paragraph: [PASSAGE\_CONTENT]\newline
    Edited:
  \end{minipage}
}
\caption{FAVA prompt for relation hallucination insertion.}
\label{tab:prompt-rela}
\end{table}

\begin{table}[b]
\centering
\fbox{%
  \begin{minipage}{\textwidth}
  \small
    For the following instruction, generate the whole output in \{lang\} language even though the instruction and input is in English.\newline
    Given a reference and a passage with possibly already inserted error tokens wrapped in \textless{}entity\textgreater{}, \textless{}relation\textgreater{}, \textless{}unverifiable\textgreater{}, \textless{}subjective\textgreater{}, or \textless{}invented\textgreater{}, insert contradictory sentence errors in the passage outside the already inserted tokens without modifying the content within already existing tokens. Wrap the inserted errors in tokens to make the passage factually incorrect. The contradictory error is defined as such:\newline
    1. contradictory sentence error (\textless{}contradictory\textgreater{}): a sentence where the entire sentence is contradicted by the given reference, meaning the sentence can be proven false due to a contradiction with information in the reference provided.\newline
    \#\#\newline
    Example 1:\newline Reference: Japan participated in World War I from 1914 to 1918 in an alliance with Entente Powers (France, the United Kingdom, Russia, the United States, Italy) against the Central Powers (Germany, Austria-Hungary, the Ottoman Empire, and Bulgaria).\newline
    Contradictory Sentence: \textless{}contradictory\textgreater{}Japan sent its army to help Germany during World War I.\textless{}/contradictory\textgreater{}\newline
    Explanation: The reference states that Japan was in an alliance against Germany, so Japan would not send its army to help Germany like the sentence states.\newline
    \#\#\newline
    Example 2:\newline Reference: Percy Jackson \& the Olympians is a series of five fantasy novels written by American author Rick Riordan.\newline Contradictory Sentence: The Harry Potter series was written by J.K Rowling\textless{}contradictory\textgreater{}, as was the Percy Jackson series\textless{}/contradictory\textgreater{}.\newline
    Explanation: The reference states that the Percy Jackson series is written by Rick Riordan and not J.K Rowling like the sentence suggests.\newline
    \#\#\newline
    Example 3:\newline Reference: As one of the busiest women in music, it'll come as no surprise that Taylor has won pretty much every award there is to win in the biz - being the proud owner of no less than 12 Grammy Awards.\newline Contradictory Sentence:\textless{}contradictory\textgreater{}Taylor Swift has never won a Grammy in her entire career since she is better known as a performer than a singer.\textless{}/contradictory\textgreater{}\newline Explanation: The reference states that Taylor Swift has won 12 Grammys and is a musician while the sentence says she has won no Grammys which contradicts the reference.\newline
    Now, insert contradictory sentences with tokens in the given passage but make sure that you don't modify anything inside any already existing \textless{}\textgreater{} error tokens at all, keep those untouched, only insert new contradictory sentences (entire sentences) with \textless{}contradictory\textgreater{}\textless{}/contradictory\textgreater{} tokens outside the already existing \textless{}\textgreater{} tags in the passage.
    \#\#\newline
    Reference: [REFERENCE\_CONTENT]\newline
    Passage: [PASSAGE\_CONTENT]\newline
    Edited:
  \end{minipage}
}
\caption{FAVA prompt for contradictory hallucination insertion.}
\label{tab:prompt-contra}
\end{table}

\begin{table}[b]
\centering
\fbox{%
  \begin{minipage}{\textwidth}
  \small
    For the following instruction, generate the whole output in \{lang\} language even though the instruction and input is in English.\newline
    Given a subject and a passage with possibly already inserted error tokens wrapped in \textless{}entity\textgreater{}, \textless{}relation\textgreater{}, \textless{}contradictory\textgreater{}, \textless{}unverifiable\textgreater{}, or \textless{}invented\textgreater{}, insert subjective sentence errors outside the already inserted tokens without modifying the content within already existing tokens. Wrap the insertions in tokens to make the passage factually incorrect. The error is defined as such:\newline
    1. subjective sentence (\textless{}subjective\textgreater{}): an entire sentence or phrase that is subjective and cannot be verified, so it should not be included.\newline
    Example 1: \textless{}subjective\textgreater{}He is the greatest soccer player ever.\textless{}/subjective\textgreater{}\newline
    Example 2: The first Harry Potter book was published in 1998 \textless{}subjective\textgreater{}and was a lot better than the rest in the series because of its use of a rich and evocative vocabulary.\textless{}/subjective\textgreater{}\newline
    Example 3: \textless{}subjective\textgreater{}Overall, Aenir is a thrilling adventure novel that takes readers on a journey through a unique and imaginative world, filling their lives with excitement.\textless{}/subjective\textgreater{}\newline
    Now, insert subjective sentence error tokens in the given passage but make sure that you don't modify anything inside any already existing \textless{}\textgreater{} error tokens at all, keep those untouched, only insert full subjective sentences or phrases with \textless{}subjective\textgreater{}\textless{}/subjective\textgreater{} tokens about the given subject outside the already existing \textless{}\textgreater{} tags in the given passage.
    \#\#\newline
    Subject: [SUBJECT\_NAME]\newline
    Passage: [PASSAGE\_CONTENT]\newline
    Edited:
  \end{minipage}
}
\caption{FAVA prompt for subjective hallucination insertion.}
\label{tab:prompt-subj}
\end{table}

\begin{table}[b]
\centering
\fbox{%
  \begin{minipage}{\textwidth}
  \small
    For the following instruction, generate the whole output in \{lang\} language even though the instruction and input is in English.\newline
    Given a reference and a passage with possibly already inserted error tokens wrapped in \textless{}entity\textgreater{}, \textless{}relation\textgreater{}, \textless{}contradictory\textgreater{}, \textless{}subjective\textgreater{}, or \textless{}invented\textgreater{}, insert unverifiable errors outside the already inserted tokens without modifying the content within already existing tokens. Wrap the insertions in tokens to make the passage factually incorrect. The error is defined as such:\newline
    1. unverifiable sentence (\textless{}unverifiable\textgreater{}): a sentence where the whole sentence or phrase is unlikely to be factually grounded although it can be true, and the sentence cannot be confirmed nor denied using the reference given or internet search, it is often something personal or private and hence cannot be confirmed.\newline
    \#\#\newline
    Unverifiable Error Example 1: \textless{}unverifiable\textgreater{}Apple is planning on releasing an instrument collection.\textless{}/unverifiable\textgreater{}\newline
    Explanation: Information about Apple’s release plans cannot be corroborated by any information online, however could be true.\newline
    \#\#\newline
    Unverifiable Error Example 2: \textless{}unverifiable\textgreater{}Selena Gomez is known to love turtles.\textless{}/unverifiable\textgreater{}\newline
    Explanation: Personal information about Selena Gomez’s opinion on turtle’s cannot be verified online, however could be true.\newline
    \#\#\newline
    Unverifiable Error Example 3: \textless{}unverifiable\textgreater{}Tom Cruise wanted to act in a Bollywood film.\textless{}/unverifiable\textgreater{}\newline
    Explanation: Personal information about Tom Cruise’s preference on acting in a Bollywood film could be true but cannot be found online.\newline
    Now, insert unverifiable error tokens in the given passage but make sure that you don't modify anything inside any already existing \textless{}\textgreater{} error tokens at all, keep those untouched, only insert unverifiable sentences or phrases with \textless{}unverifiable\textgreater{}\textless{}/unverifiable\textgreater{} tokens outside the already existing \textless{}\textgreater{} tags in the given passage. Remember, unverifiable sentences seem like they are true but cannot be confirmed or denied.
    \#\#\newline
    Reference: [REFERENCE\_CONTENT]\newline
    Passage: [PASSAGE\_CONTENT]\newline
    Edited:
  \end{minipage}
}
\caption{FAVA prompt for unverifiable hallucination insertion.}
\label{tab:prompt-unv}
\end{table}

\begin{table}[b]
\centering
\fbox{%
  \begin{minipage}{\textwidth}
  \small
    For the following instruction, generate the whole output in \{lang\} language even though the instruction and input is in English.\newline
    Given a subject and a passage with possibly already inserted error tokens wrapped in \textless{}entity\textgreater{}, \textless{}relation\textgreater{}, \textless{}contradictory\textgreater{}, \textless{}unverifiable\textgreater{}, or \textless{}subjective\textgreater{}, insert invented info sentence errors outside the already inserted tokens without modifying the content within already existing tokens. Wrap the errors in tokens to make the passage factually incorrect. The error is defined as such:\newline
    1. invented info error (\textless{} invented \textgreater{}): these errors refer to entities that are not known or do not exist. this does not include fictional characters in books or movies. invented info errors include phrases or sentences which have unknown entities or misleading information.\newline
    \#\#\newline
    Invented Info Example 1: \textless{}invented\textgreater{}Kansas City has a large population of the Yuman Tribe.\textless{}/invented\textgreater{}\newline
    Explanation: Yuman tribe is not an actual tribe, they are a invented entity.\newline
    \#\#\newline
    Invented Info Example 2: Joel Embiid is a Cameroonian professional basketball player for the Philadelphia 76ers , and he was awarded the Kia NBA MVP Trophy in 2023 \textless{}invented\textgreater{}and received the Shaquille O’Neal trophy for being the fastest runner this season.\textless{}/invented\textgreater{}\newline
    Explanation: There is no trophy named the Shaquille O’Neal trophy in NBA and so it is not possible for Joel Embiid to have won it. Also, there is no award for being the fastest runner in the NBA. Both are invented.\newline
    \#\#\newline
    Invented Info Example 3: \textless{}invented\textgreater{}Andrew Ng’s area of Sentiment-Infused Language Generation (SILG) which explores the influence of sentiment analysis on the generation of human-like language.\textless{}/invented\textgreater{}\newline
    Explanation: There is no field of research named Semantic compositionality analysis, so it is a invented entity.\newline
    Now, insert invented information error tokens about the subject in the given passage but make sure that you don't modify anything inside any already existing \textless{}\textgreater{} error tokens at all, keep those untouched, only add fictional sentence errors with \textless{}invented\textgreater{}\textless{}/invented\textgreater{} tokens outside the already existing \textless{}\textgreater{} tags in the given passage. Also avoid inserting errors before the first sentence. Also make sure you tag each edit with \textless{}invented\textgreater{}\textless{}/invented\textgreater{} tags.
    \#\#\newline
    Subject: [SUBJECT\_NAME]\newline
    Passage: [PASSAGE\_CONTENT]\newline
    Edited:
  \end{minipage}
}
\caption{FAVA prompt for invented hallucination insertion.}
\label{tab:prompt-inv}
\end{table}

\end{document}